\documentclass[a4paper, twoside, 12pt]{article}

\usepackage[noblocks]{authblk} 
\usepackage[numbers]{natbib}
\usepackage{graphicx}
\usepackage{fancyhdr}
\usepackage[colorlinks=true, linkcolor=MidnightBlue, urlcolor=MidnightBlue, citecolor=PineGreen]{hyperref}
\usepackage[tableposition=top, justification=justified, textfont=it]{caption}
\usepackage{titlesec}
\usepackage{multicol}
\usepackage{afterpage}
\usepackage{lipsum}
\usepackage[dvipsnames, x11names]{xcolor}
\usepackage[inner=4cm, outer=4cm, top=2.8cm, bottom=2.8cm, marginparsep=0cm, marginparwidth=0cm, includehead, includefoot]{geometry}
\usepackage{setspace}
\usepackage{array}
\usepackage{bm}
\usepackage{tikz}
\usepackage{float}
\usepackage[framemethod]{mdframed}
\usepackage{amsmath}
\usepackage{amssymb}

\titleformat{\section}{\Large \bfseries \centering \scshape}{\thesection.}{0.3em}{}[{\titlerule[0.5pt]}]

\definecolor{shadecolor}{RGB}{230,230,230}
\newcommand{\mybox}[1]{\par\noindent\colorbox{shadecolor}
{\parbox{\dimexpr\textwidth-2\fboxsep\relax}{#1}}}

\titleformat{\subsection}{\large \bfseries \mybox}{\thesubsection}{1em}{}

\titleformat{\subsubsection}{\itshape}{\thesubsubsection.}{0.3em}{}

\renewenvironment{abstract}
{\vskip 2.5ex {\large\bf\noindent Abstract}\vspace{0.7ex} \\ %
  \bgroup\noindent\ignorespaces}%
{\par\egroup\vskip 2.5ex}

\newenvironment{keywords}
{\bgroup\leftskip 20pt\rightskip 20pt \small\noindent{\bf Keywords:} }%
{\par\egroup\vskip 10ex}

\fancypagestyle{firstpage}{%
  
  \fancyhead[]{}
  \fancyfoot[C]{}
  \fancyfoot[L]{\footnotesize To appear in \\
  O. Colliot (Ed.), \textit{Machine Learning for Brain Disorders}, Springer\\
  }

}

\fancypagestyle{otherpages}{%
  
  \fancyhead[LE]{\thepage}
  \fancyhead[RE]{\runningauthor}
  \fancyhead[LO]{\runningheadtitle}
  \fancyhead[RO]{\thepage}
  \fancyfoot[L]{\footnotesize  \textit{Machine Learning for Brain Disorders}, Chapter~\chapternumber}
  \fancyfoot[C]{}
}

\makeatletter
\renewcommand{\maketitle}{\bgroup\setlength{\parindent}{0pt}

\begin{flushright}
  \color{MidnightBlue}
  \textbf{\LARGE Chapter~\chapternumber}
\end{flushright}

\vspace{0.3in}

\begin{flushleft}
    \setstretch{2.0} 
    \textbf{\color{MidnightBlue}\huge\@title}
\end{flushleft}

\vspace{0.15in}

\begin{flushleft}
    \textbf{\bfseries \large\@author}
\end{flushleft}\egroup
}

\makeatother

\DeclareCaptionLabelFormat{labelformat}{\color{MidnightBlue}\bfseries #1 #2}
\renewcommand{\bibpreamble}{\scriptsize \begin{multicols}{2}}
\renewcommand{\bibpostamble}{\end{multicols}}

\mdfsetup{skipabove=\topskip, skipbelow=\topskip}

\newcounter{nicebox}
\newenvironment{nicebox}[1][]{%
    \refstepcounter{nicebox}%
    \ifstrempty{#1}%
    {\mdfsetup{%
        frametitle={%
            \tikz[baseline=(current bounding box.east),outer sep=0pt]
            \node[anchor=east,rectangle,fill=blue!20]
            {\strut Theorem~\thetheo};}}
    }%
    {\mdfsetup{%
        frametitle={%
            \tikz[baseline=(current bounding box.east),outer sep=0pt]
            \node[anchor=east,rectangle,fill=blue!20]
            {\strut Box~\thenicebox:~#1};}}%
    }%
    \mdfsetup{innertopmargin=10pt,linecolor=blue!20, linewidth=2pt,topline=true, frametitleaboveskip=\dimexpr-\ht\strutbox\relax,}
    \begin{mdframed}[]\relax%
    }{\end{mdframed}}

\newfloat{floatbox}{thp}{lop}
\floatname{floatbox}{Box}

\DeclareMathOperator*{\argmax}{arg\,max}
\DeclareMathOperator*{\argmin}{arg\,min}

\DeclareMathAlphabet{\mathsfit}{\encodingdefault}{\sfdefault}{m}{sl}
\SetMathAlphabet{\mathsfit}{bold}{\encodingdefault}{\sfdefault}{bx}{n}

\usepackage{afterpage}
\usepackage[algoruled, vlined]{algorithm2e}
\usepackage{amsfonts}
\usepackage{caption}
\usepackage[inline]{enumitem}
\usepackage{subcaption}
\usepackage{tikz}
\usetikzlibrary{fit, positioning, arrows.meta, positioning, calc, decorations.pathreplacing, matrix}
\usepackage{longtable}

\makeatletter
\tikzset{
    from/.style args={#1 to #2}{
        above right={0cm of #1},
        /utils/exec=\pgfpointdiff
            {\tikz@scan@one@point\pgfutil@firstofone(#1)\relax}
            {\tikz@scan@one@point\pgfutil@firstofone(#2)\relax},
        minimum width/.expanded=\the\pgf@x,
        minimum height/.expanded=\the\pgf@y}}
\makeatother

\definecolor{C0}{RGB}{31, 119, 180}
\definecolor{C1}{RGB}{255, 127, 14}
\definecolor{C3}{RGB}{214, 39, 40}

\allowdisplaybreaks

\begin{document}


\newcommand{\runningauthor}{Faouzi and Colliot} 

\newcommand{\runningheadtitle}{Classic machine learning methods}

\newcommand{\chapternumber}{2}

\newcommand{\emailaddress}{johann.faouzi@gmail.com}

\title{Classic machine learning methods} 

\author[*1]{Johann Faouzi}
\author[2]{Olivier Colliot}
\affil[1]{CREST, ENSAI, Campus de Ker-Lann, 51 Rue Blaise Pascal, BP 37203 -- 35172 Bruz Cedex, France}
\affil[2]{Sorbonne Universit\'e, Institut du Cerveau - Paris Brain Institute - ICM, CNRS, Inria, Inserm, AP-HP, H\^opital de la Piti\'e-Salp\^etri\`ere, F-75013, Paris, France}
\affil[*]{Corresponding author: e-mail address: \href{mailto:\emailaddress}{\emailaddress}}

\maketitle

\afterpage{\aftergroup\restoregeometry}
\pagestyle{otherpages}

\begin{abstract}
In this chapter, we present the main classic machine learning methodss. A large part of the chapter is devoted to supervised learning techniques for classification and regression, including nearest-neighbor methods, linear and logistic regressions, support vector machines and tree-based algorithms. We also describe the problem of overfitting as well as strategies to overcome it. We finally provide a brief overview of unsupervised learning methods, namely for clustering and dimensionality reduction. The chapter does not cover neural networks and deep learning as these will be presented in Chapters 3, 4, 5 and 6.
\end{abstract}

\begin{keywords}
    machine learning, classification, regression, clustering, dimensionality reduction
\end{keywords}

\section{Introduction}

This chapter presents the main classic machine learning (ML) methods. There is a focus on supervised learning methods for classification and regression, but we also describe some unsupervised approaches. The chapter is meant to be readable by someone with no background in machine learning. It is nevertheless necessary to have some basic notions of linear algebra, probabilities and statistics. If this is not the case, we refer the reader to chapters 2 and 3 of \citep{goodfellow2016deep}.

The rest of this chapter is organized as follows. Rather than grouping methods by categories (for instance classification or regression methods), we chose to present methods by increasing order of complexity. We first provide the notations in \autoref{sec:notations}. We then describe a very intuitive family of methods, that of nearest neighbors (\autoref{sec:nearest}). We continue with linear regression (\autoref{sec:linear}) and logistic regression (\autoref{sec:logistic}), the later being a classification technique. We subsequently introduce the problem of overfitting (\autoref{sec:overfitting}) as well as strategies to mitigate it (\autoref{sec:penalized}). \autoref{sec:support_vector_machine} describes support vector machines (SVM). \autoref{sec:multiclass} explains how binary classification methods can be extended to a multi-class setting. We then describe methods which are specifically adapted to the case of normal distributions (\autoref{sec:decision_functions_normal_distributions}). Decision trees and random forests are described in \autoref{sec:tree}. We then briefly describe some unsupervised learning techniques, namely for clustering (\autoref{sec:clustering}) and dimensionality reduction (\autoref{sec:dimensionality_reduction}). The chapter ends with a description of kernel methods which can be used to extend linear techniques to non-linear cases (\autoref{sec:kernel_methods}).
\autoref{box:algorithms} summarizes the methods presented in this chapter, grouped by categories and then sorted in order of appearance.

\begin{floatbox}[hbtp]
    \begin{nicebox}[Main classic ML methods]
        \label{box:algorithms}
        \begin{itemize}[leftmargin=2mm]
            \item \textbf{Supervised learning}
            \begin{itemize}
                \item \textbf{Classification}: nearest neighbors, logistic regression, support vector machine (SVM), naive Bayes, linear discriminant analysis (LDA), quadratic discriminant analysis, tree-based models (decision tree, random forest, extremely randomized trees).
                \item \textbf{Regression}: nearest-neighbors, linear regression, support vector machine regression, tree-based models (decision tree, random forest, extremely randomized trees), kernel ridge regression.
            \end{itemize}
            \item \textbf{Unsupervised learning}
            \begin{itemize}
                \item \textbf{Clustering}: $k$-means, Gaussian mixture model.
                \item \textbf{Dimensionality reduction}: principal component analysis (PCA), linear discriminant analysis (LDA), kernel principal component analysis.
            \end{itemize}
        \end{itemize}
    \end{nicebox}
\end{floatbox}

\section{Notations}
\label{sec:notations}

Let $n$ be the number of samples and $p$ be the number of features. An input sample is thus a $p$-dimensional vector:
$$\bm{x}=\left[\begin{array}{c}{x_{1}} \\ {\vdots} \\ {x_{p}}\end{array}\right]$$
An output sample is denoted by $y$. Thus, a sample is $(\bm{x},y)$.
The dataset of $n$ samples can then be summarized as an $n \times p$ matrix $\bm{X}$ representing the input data and an $n$-dimensional vector $\bm{y}$ representing the target data:
$$
\bm{X} = 
\begin{bmatrix} \bm{x}^{(1)} \\ \vdots \\ \bm{x}^{(n)} \end{bmatrix} =
\begin{bmatrix}
x_{1}^{(1)} & \ldots & x_{p}^{(1)} \\
\vdots & \ddots & \vdots \\
x_{1}^{(n)} & \ldots & x_{p}^{(n)}
\end{bmatrix}
\qquad , \qquad
\bm{y} = \begin{bmatrix} y_1 \\ \vdots \\ y_n \end{bmatrix}
$$
The input space is denoted by $\mathcal{I}$ and the set of training samples is denoted by $\mathcal{X}$.

In the case of regression, $y$ is a real number.
In the case of classification, $y$ is a single label.
More precisely, $y$ can only take one of a finite set of values called labels.
The set of possible classes (i.e., labels) is denoted by $\mathcal{C} = \{ \mathcal{C}_1, \ldots, \mathcal{C}_q \}$, with $q$ being the number of classes.
As the values of the classes are not meaningful, when there are only two classes, the classes are often called the positive and negative classes.
In this case and also for mathematical reasons, without loss of generality, we assume the values of the classes to be $+1$ and $-1$.


\section{Nearest-neighbor methods}
\label{sec:nearest}

One of the most intuitive approaches to machine learning is nearest neighbors. It is based on the following intuition: for a given input, its corresponding output is likely to be similar to the outputs of similar inputs.
A real-life metaphor would be that, if a subject has similar characteristics than other subjects who were diagnosed with a given disease, then this subject is likely to also be suffering from this disease.

More formally, nearest-neighbor methods use the training samples from the neighborhood of a given point $\bm{x}$, denoted by $N(\bm{x})$, to perform prediction \cite{murphy_machine_2012}.

For regression tasks, the prediction is computed as a weighted mean of the target values in $N(\bm{x})$:
$$
\hat{y} = \sum_{\bm{x}^{(i)} \in N(\bm{x})} w_i^{(\bm{x})} y^{(i)}
$$
where $w_i^{(\bm{x})}$ is the weight associated to $\bm{x}^{(i)}$ to predict the output of $\bm{x}$, with $w_i^{(\bm{x})} \geq 0 \; \forall i$ and $\sum_{i} w_i^{(\bm{x})} = 1$.

For classification tasks, the predicted label corresponds to the label with the largest weighted sum of occurrences of each label:
$$
\hat{y} = \argmax_{\mathcal{C}} \sum_{\bm{x}^{(i)} \in N(\bm{x})} w_i^{(\bm{x})} \bm{1}_{y^{(i)} = \mathcal{C}_k}
$$

A key parameter of nearest-neighbor methods is the \emph{metric}, denoted by $d$, that is a mathematical function that defines dissimilarity.
The metric is used to define the neighborhood of any point and can also be used to compute the weights.

\subsection{Metrics}

Many metrics have been defined for various types of input data such as vectors of real numbers, integers or booleans.
Among these different types, vectors of real numbers are one of the most common types of input data, for which the most commonly used metric is the Euclidean distance, defined as:
$$
\forall \bm{x}, \bm{x}' \in \mathcal{I},\, \Vert \bm{x} - \bm{x}' \Vert_2
= \sqrt{ \sum_{j=1}^p (x_j - x'_j)^2 }
$$
The Euclidean distance is sometimes referred to as the ``ordinary'' distance since it is the one based on the Pythagorean theorem and that everyone uses in their everyday lives.

\subsection{Neighborhood}

\begin{figure}
    \centering
    \includegraphics[width=1.\textwidth]{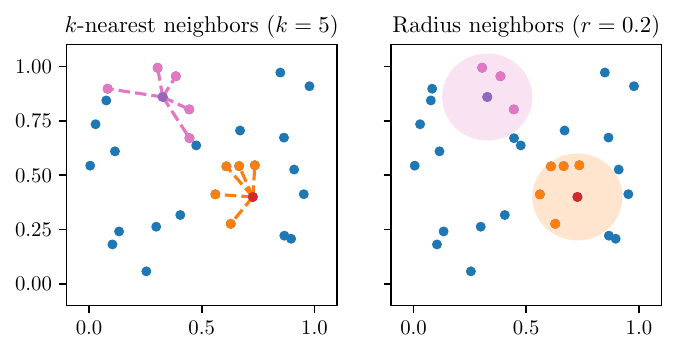}
    \caption{Different definitions of the neighborhood. On the left, the neighborhood of a given point is the set of its 5 nearest neighbors. On the right, the neighborhood of a given point is the set of points whose dissimilarity is lower than the radius. For a given input, its neighborhood may be different depending on the definition used. The Euclidean distance is used as the metric in both examples.}
    \label{fig:neighborhood_types}
\end{figure}

The two most common definitions of the neighborhood rely on either the number of neighbors or the radius around the given point.
\autoref{fig:neighborhood_types} illustrates the differences between both definitions.

The $k$-nearest neighbor method defines the neighborhood of a given point $\bm{x}$ as the set of the $k$ closest points to $\bm{x}$:
$$
N(\bm{x}) = \{ \bm{x}^{(i)} \}_{i=1}^k \quad \text{with}\quad d(\bm{x}, \bm{x}^{(1)}) \leq \ldots \leq d(\bm{x}, \bm{x}^{(n)})
$$

The radius neighbor method defines the neighborhood of a given point $\bm{x}$ as the set of points whose dissimilarity to $\bm{x}$ is smaller than the given radius, denoted by $r$:
$$
N(\bm{x}) = \{ \bm{x}^{(i)} \in \mathcal{X} \,|\, d(\bm{x}, \bm{x}^{(i)}) < r \}
$$

\subsection{Weights}

\begin{figure}
    \centering
    \includegraphics[width=1.\textwidth]{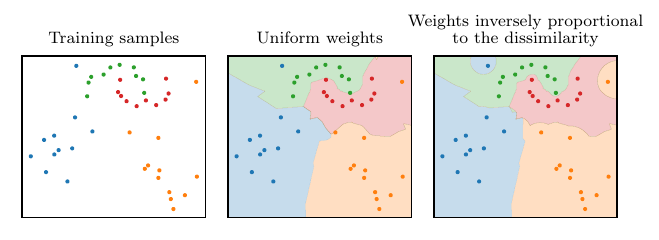}
    \caption{Impact of the definition of the weights on the prediction function of a 3-nearest neighbor classification model. When the weights are inversely proportional to the dissimilarity, the classifier is more subject to outliers since the predictions in the close neighborhood of any input is mostly dedicated by the label of this input, independently of the number of neighbors used. With uniform weights, the prediction function tends to be smoother.}
    \label{fig:nearest_neighbor_classification_weights}
\end{figure}

The two most common approaches to compute the weights are to use:
\begin{itemize}
    \item uniform weights (all the weights are equal):
    $$\forall i,\, w_i^{(\bm{x})} = \frac{1}{\vert N(\bm{x}) \vert}$$
    \item weights inversely proportional to the dissimilarity:
    $$
    \forall i,\, w_i^{(\bm{x})}
    = \frac{\frac{1}{d(\bm{x}^{(i)}, \bm{x})}}{\sum_{j} \frac{1}{d(\bm{x}^{(j)}, \bm{x})}}
    = \frac{1}{d(\bm{x}^{(i)}, \bm{x}) \sum_{j} \frac{1}{d(\bm{x}^{(j)}, \bm{x})}}
    $$
\end{itemize}
With uniform weights, every point in the neighborhood equally contributes to the prediction.
With weights inversely proportional to the dissimilarity, closer points contribute more to the prediction than further points.
\autoref{fig:nearest_neighbor_classification_weights} illustrates the different decision functions obtained with uniform weights and weights inversely proportional to the dissimilarity for a $3$-nearest neighbor classification model.

\subsection{Neighbor search}

The brute-force method to compute the neighborhood for $n$ points with $p$ features is to compute the metric for each pair of inputs, which has a $\mathcal{O}(n^2p)$ algorithmic complexity (assuming that evaluating the metric for a pair of inputs has a complexity of $\mathcal{O}(p)$, which is the case for most metrics).
However, it is possible to decrease this algorithmic complexity if the metric is a \emph{distance}, that is if the metric $d$ satisfies the following properties:
\begin{enumerate}
    \item Non-negativity: $\forall \bm{a}, \bm{b}, \, d(\bm{a}, \bm{b}) \geq 0$
    \item Identity: $\forall \bm{a}, \bm{b}, \, d(\bm{a}, \bm{b}) = 0$ if and only if $\bm{a} = \bm{b}$
    \item Symmetry: $\forall \bm{a}, \bm{b}, \, d(\bm{a}, \bm{b}) = d(\bm{b}, \bm{a})$
    \item Triangle Inequality: $\forall \bm{a}, \bm{b}, \bm{c}, \, d(\bm{a}, \bm{b}) + d(\bm{b}, \bm{c}) \geq d(\bm{a}, \bm{c})$
\end{enumerate}
The key property is the \emph{triangle inequality}, which has a simple interpretation: the shortest path between two points is a straight line.
Mathematically, if $\bm{a}$ is far from $\bm{c}$ and $\bm{c}$ is close to $\bm{b}$ (i.e., $d(\bm{a}, \bm{c})$ is large and $d(\bm{b}, \bm{c})$ is small), then $\bm{a}$ is far from $\bm{b}$ (i.e., $d(\bm{a}, \bm{b})$ is large). This is obtained by rewriting the triangle inequality as follows:
$$
\forall \bm{a}, \bm{b}, \bm{c}, \,  d(\bm{a}, \bm{b}) \geq d(\bm{a}, \bm{c}) - d(\bm{b}, \bm{c})
$$
This means that it is not necessary to compute $d(\bm{a}, \bm{b})$ in this case.
Therefore, the computational cost of a nearest neighbors search can be reduced to $\mathcal{O}(n \log(n)p)$ or better, which is a substantial improvement over the brute-force method for large $n$.
Two popular methods that take advantage of this property are the \emph{$K$-dimensional tree} structure \citep{bentley_multidimensional_1975} and the \emph{ball tree} structure \citep{omohundro_five_1989}.

\section{Linear regression}
\label{sec:linear}
Linear regression is a regression model that linearly combines the features.
Each feature is associated with a coefficient that represents the relative weight of this feature compared to the other features.
A real-life metaphor would be to see the coefficients as the ingredients of a recipe: the key is to find the best balance (i.e., proportions) between all the ingredients in order to make the best cake.

Mathematically, a linear model is a model that linearly combines the features \cite{bishop_pattern_2006}:
$$
f(\bm{x}) = w_0 + \sum_{j=1}^p w_j x_j
$$
A common notation consists in including a $1$ in $\bm{x}$ so that $f(\bm{x})$ can be written as the \emph{dot product} between the vector $\bm{x}$ and the vector $\bm{w}$:
$$
f(\bm{x}) = w_0 \times 1 + \sum_{j=1}^p w_j x_j = \bm{x}^\top \bm{w}
$$
where the vector $\bm{w}$ consists of:
\begin{itemize}
    \item the intercept (also known as bias) $w_0$, and
    \item the coefficients $(w_1, \ldots, w_p)$, where each coefficient $w_j$ is associated with the corresponding feature $x_j$.
\end{itemize}

In the case of linear regression, $f(\bm{x})$ is the predicted output:
$$
\hat{y} = f(\bm{x}) = \bm{x}^\top \bm{w}
$$
There are several methods to estimate the $\bm{w}$ coefficients.
In this section, we present the oldest one which is known as \emph{ordinary least squares regression}.

\begin{figure}[hbtp]
    \centering
    \includegraphics[width=1.\textwidth]{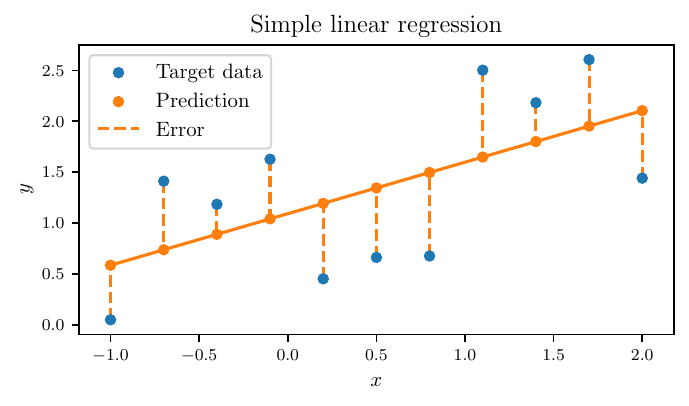}
    \caption{Ordinary least squares regression. The coefficients (that is the intercept and the slope with a single predictor) are estimated by minimizing the sum of the squared errors.}
    \label{fig:linear_regression}
\end{figure}

In the case of ordinary least squares regression, the cost function $J$ is the sum of the squared errors on the training data (see \autoref{fig:linear_regression}):
$$
J(\bm{w}) = \sum_{i=1}^n \left( y^{(i)} - \hat{y}^{(i)} \right)^2 = \sum_{i=1}^n \left( y^{(i)} - \bm{x}^{(i)\top} \bm{w} \right)^2 = \Vert \bm{y} - \bm{X} \bm{w} \Vert_2^2
$$
One wants to find the optimal parameters $\bm{w}^\star$ that minimize the cost function:
$$
\bm{w}^\star= \argmin_{\bm{w}} J(\bm{w})
$$
This optimization problem is \emph{convex}, implying that any local minimum is a global minimum, and \emph{differentiable}, implying that every local minimum has a null gradient.
One therefore aims to find null gradients of the cost function:
\begin{align*}
    \nabla_{\bm{w}^\star} J &= 0 \\
    \implies 2 \bm{X}^\top \bm{X} \bm{w}^\star - 2 \bm{X}^\top \bm{y} &= 0 \\
    \implies \bm{X}^\top \bm{X} \bm{w}^\star &= \bm{X}^\top \bm{y} \\
    \implies \bm{w}^\star &= \left(\bm{X}^\top \bm{X} \right)^{-1} \bm{X}^\top \bm{y}
\end{align*}

Ordinary least square regression is one of the few machine learning optimization problem for which there exists a \emph{closed formula}, i.e. the optimal solution can be computed using a finite number of standard operations such as addition, multiplication and evaluations of well-known functions.

\begin{floatbox}[hbtp]
    \begin{nicebox}[Linear regression]
        \label{box:linear_regression}
        \begin{itemize}[leftmargin=2mm]
            \item \textbf{Main idea}: best hyperplane (i.e., line when $p=1$, plane when $p=2$) mapping the inputs and to the outputs. 
            \item \textbf{Mathematical formulation}: linear relationship between the predicted output $\hat{y}$ and the input $\bm{x}$ that minimizes the sum of squared errors:
            $$
            \hat{y} = w_0^\star + \sum_{j=1}^n w_j^\star x_j \quad \textnormal{with} \quad \bm{w}^\star = \argmin_{\bm{w}} \sum_{i=1}^n \left( y^{(i)} - \bm{x}^{(i)\top} \bm{w} \right)^2
            $$
            \item \textbf{Regularization}: can be penalized to avoid overfitting (ridge), to perform feature selection (lasso), or both (elastic-net). See \autoref{sec:penalized}. 
        \end{itemize}
    \end{nicebox}
\end{floatbox}

\section{Logistic regression}
\label{sec:logistic}
Intuitively, linear regression consists in finding the line that best fits the data: the true output should be as close to the line as possible.
For binary classification, one wants the line to separate both classes as well as possible: the samples from one class should all be in one subspace and the samples from the other class should all be in the other subspace, with the inputs being as far as possible from the line.

\begin{figure}
    \centering
    \includegraphics[width=0.6\textwidth]{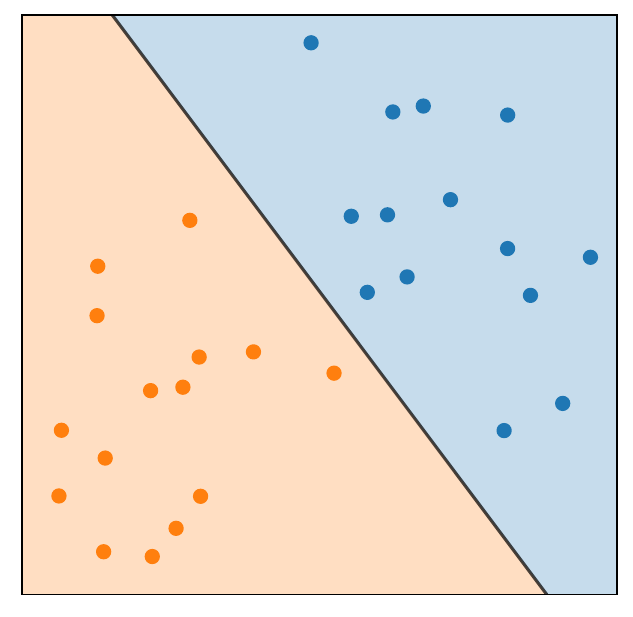}
    \caption[Decision function of a logistic regression model]
    {Decision function of a logistic regression model.
    A logistic regression is a linear model, that is its decision function is linear.
    In the two-dimensional case, it separates a plane with a line.}
    \label{fig:logistic_regression}
\end{figure}

Mathematically, for binary classification tasks, a linear model is defined by an hyperplane splitting the input space into two subspaces such that each subspace is characteristic of one class.
For instance, a line splits a plane into two subspaces in the two-dimensional case, while a plane splits a three-dimensional space into two subspaces.
A hyperplane is defined by a vector $\bm{w} = (w_0, w_1, \ldots, w_p)$ and $f(\bm{x}) = \bm{x}^\top \bm{w}$ corresponds to the \emph{signed distance} between the input $\bm{x}$ and the hyperplane $\bm{w}$: in one subspace, the distance with any input is always positive, whereas in the other subspace, the distance with any input is always negative.
\autoref{fig:logistic_regression} illustrates the decision function in the two-dimensional case where both classes are linearly separable.

The sign of the signed distance corresponds to the decision function of a linear binary classification model:
$$
\hat{y} = \text{sign}(f(\bm{x})) = \begin{cases} +1 & \text{if $f(\bm{x}) > 0$}\\ -1 & \text{if $f(\bm{x}) < 0$} \end{cases}
$$

The logistic regression model is a probabilistic linear model that transforms the signed distance to the hyperplane into a probability using the sigmoid function \citep{hastie_elements_2009}, denoted by $\sigma(u)=\frac{1}{1 + \exp\left( - u \right)}$.

Consider the linear model:
$$
    f(\bm{x}) = \bm{x}^\top \bm{w} = w_0 + \sum_{i=j}^p w_j x_j
$$
Then the probability of belonging to the positive class is:
$$
    P \left( \textnormal{y}=+1 | \mathbf{x} = \bm{x} \right) = \sigma(f(\bm{x})) = \frac{1}{1 + \exp\left( - f(\bm{x}) \right)}
$$
and that of belonging to the negative class is:
\begin{align*}
    P \left( \textnormal{y}=-1 | \mathbf{x} = \bm{x} \right) &= 1 - P \left( \textnormal{y}=+1 | \mathbf{x} = \bm{x} \right) \\
    &= \frac{\exp\left( -f(\bm{x}) \right)}{1 + \exp\left( -f(\bm{x}) \right)} \\
    &= \frac{1}{1 + \exp\left( f(\bm{x}) \right)} \\
    P \left( \textnormal{y}=-1 | \mathbf{x} = \bm{x} \right) &= \sigma(-f(\bm{x}))
\end{align*}

By applying the inverse of the sigmoid function, which is known as the logit function, one can see that the logarithm of the \emph{odds ratio} is modeled as a linear combination of the features:
$$
    \log\left( \frac{P \left( \textnormal{y}=+1 | \mathbf{x} = \bm{x} \right)}{P \left( \textnormal{y}=-1 | \mathbf{x} = \bm{x} \right)}  \right) = 
    \log\left( \frac{P \left( \textnormal{y}=+1 | \mathbf{x} = \bm{x} \right)}{1 - P \left( \textnormal{y}=+1 | \mathbf{x} = \bm{x} \right)} \right) = 
    f(\bm{x})
$$

\begin{floatbox}[t]
    \begin{nicebox}[Logistic regression]
        \label{box:logistic_regression}
        \begin{itemize}[leftmargin=2mm]
            \item \textbf{Main idea}: best hyperplane (i.e., line) that separates two classes. 
            \item \textbf{Mathematical formulation}: the signed distance to the hyperplane is mapped into the probability to belong to the positive class using the sigmoid function:
            \begin{gather*}
                f(\bm{x}) = w_0 + \sum_{j=1}^n w_j x_j \\
                P \left( \textnormal{y}=+1 | \mathbf{x} = \bm{x} \right) = \sigma(f(\bm{x})) = \frac{1}{1 + \exp\left( - f(\bm{x}) \right)}
            \end{gather*}
            \item \textbf{Estimation}: likelihood maximization.
            \item \textbf{Regularization}: can be penalized to avoid overfitting ($\ell_2$ penalty), to perform feature selection ($\ell_1$ penalty), or both (elastic-net penalty). 
        \end{itemize}
    \end{nicebox}
\end{floatbox}

The $\bm{w}$ coefficients are estimated by maximizing the \emph{likelihood} function, that is the function measuring the goodness of fit of the model to the training data:
$$
L(\bm{w}) = \prod_{i=1}^n P \left( \textnormal{y}=y^{(i)} | \mathbf{x} = \bm{x}^{(i)} ; \bm{w} \right)
$$
For computational reasons, it is easier to maximize the \emph{log-likelihood}, which is simply the logarithm of the likelihood:
\begin{align*}
    \log(L(\bm{w}))
    &= \sum_{i=1}^n \log \left( P \left( \textnormal{y}=y^{(i)} | \mathbf{x} = \bm{x}^{(i)} ; \bm{w} \right) \right) \\
    &= \sum_{i=1}^n \log \left( \sigma\left( y^{(i)} f(\bm{x}^{(i)} ; \bm{w}) \right) \right) \\
    &= \sum_{i=1}^n -\log\left( 1 + \exp \left(y^{(i)} \bm{x}^{(i)\top} \bm{w} \right) \right) \\
    \log(L(\bm{w}))
    &= - \sum_{i=1}^n \log\left( 1 + \exp \left(y^{(i)} \bm{x}^{(i)\top} \bm{w} \right) \right)
\end{align*}
Finally, we can rewrite this maximization problem as a minimization problem by noticing that $\max_{\bm{w}} \log(L(\bm{w})) = - \min_{\bm{w}} - \log(L(\bm{w}))$:
$$
\max_{\bm{w}} \log(L(\bm{w})) = - \min_{\bm{w}} \sum_{i=1}^n \log\left( 1 + \exp \left(y^{(i)} \bm{x}^{(i)\top} \bm{w} \right) \right)
$$
We can see that the $\bm{w}$ coefficients that maximize the likelihood are also the coefficients that minimize the sum of the \emph{logistic loss} values, with the logistic loss being defined as:
$$
\ell_{\text{logistic}}(y, f(\bm{x})) = \log\left( 1 + \exp \left(y f(\bm{x}) \right) \right) / \log(2)
$$
Unlike for linear regression, there is no closed formula for this minimization.
One thus needs to use an optimization method such as for instance gradient descent which was presented in Section 3 of Chapter 1.
In practice, more sophisticated approaches such as quasi-Newton methods and variants of stochastic gradient descent are often used.

\section{Overfitting and regularization}
\label{sec:overfitting}
The original formulations of ordinary least square regression and logistic regression are \emph{unregularized} models, that is the model is trained to fit the training data as much as possible.
Let us consider a real-life example as it is very similar to human learning.
If a person learns by heart the content of a book, they are able to solve the exercises in the book, but unable to apply the theoretical concepts to new exercises or real-life situations.
If a person only quickly reads through the book, they are probably unable to solve neither the exercises in the book nor new exercises.

The corresponding concepts are known as \emph{overfitting} and \emph{underfitting} in machine learning.
Overfitting occurs when a model fits too well the training data and generalizes poorly to new data.
Oppositely, underfitting occurs when a model does not capture well enough the characteristics of the training data, thus also generalizes poorly to new data.

Overfitting and underfitting are related to frequently used terms in machine learning: \emph{bias} and \emph{variance}.
Bias is defined as the expected (i.e., mean) difference between the true output and the predicted output.
Variance is defined as the variability of the predicted output.
For instance, let us consider a model predicting the age of a person from a picture.
If the model \emph{always} underestimates or overestimates the age, then the model is biased.
If the model makes \emph{both large and small errors}, then the model has a high variance.

\begin{figure}
    \centering
    \includegraphics[width=.8\textwidth]{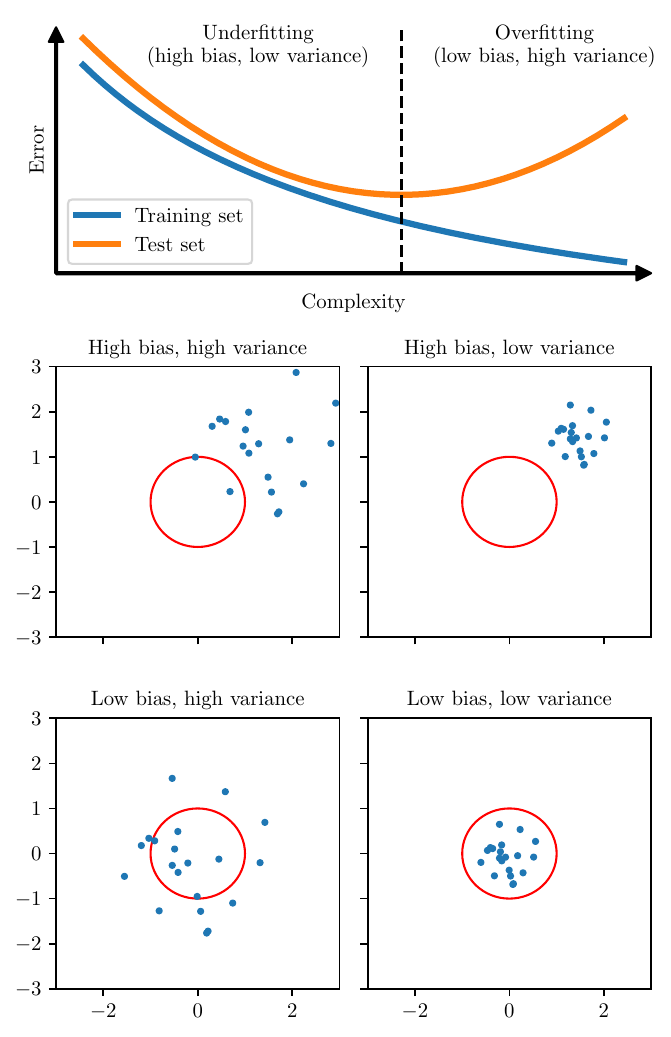}
    \caption{Illustration of underfitting and overfitting.
    Underfitting occurs when a model is too simple and does not capture well enough the characteristics of the training data, leading to high bias and low variance.
    Oppositely, overfitting occurs when a model is too complex and learns the noise in the training data, leading to low bias and high variance.}
    \label{fig:underfitting_overfitting_bias_variance}
\end{figure}

Ideally, one would like to have a model with a small bias and a small variance.
However, the bias of a model tends to increase when decreasing its variance, and the variance of the model tends to increase when decreasing its bias.
This phenomenon is known as the \emph{bias-variance trade-off}.
\autoref{fig:underfitting_overfitting_bias_variance} illustrates this phenomenon.
One can also notice it by computing the squared error between the true output $y$ (fixed) and the predicted output $\hat{\textnormal{y}}$ (random variable): its expected value is the sum of the squared bias of $\hat{\textnormal{y}}$ and the variance of $\hat{\textnormal{y}}$:

\begin{align*}
    \mathbb{E} \left[ (y  - \hat{\textnormal{y} })^2 \right]
    &= \mathbb{E} \left[ y ^2 - 2 y  \hat{\textnormal{y} } + \hat{\textnormal{y} }^2 \right] \\
    &= y ^2 - 2 y  \mathbb{E} \left[ \hat{\textnormal{y} } \right] + \mathbb{E} \left[ \hat{\textnormal{y} }^2 \right] \\
    &= y^2 - 2 y  \mathbb{E} \left[ \hat{\textnormal{y} } \right] + \mathbb{E} \left[ \hat{\textnormal{y} }^2 \right] + \mathbb{E} \left[ \hat{\textnormal{y} } \right]^2 - \mathbb{E} \left[ \hat{\textnormal{y} } \right]^2  \\
    &= \left( \mathbb{E} \left[ \hat{\textnormal{y} } \right] - y \right)^2 + \mathbb{E} \left[ \hat{\textnormal{y} }^2 \right] - \mathbb{E} \left[ \hat{\textnormal{y} } \right]^2  \\
    &= \left( \mathbb{E} \left[ \hat{\textnormal{y} } \right] - y \right)^2 + \mathbb{E} \left[ \hat{\textnormal{y} }^2 - \mathbb{E} \left[ \hat{\textnormal{y} } \right]^2 \right] \\
    &= \left( \mathbb{E} \left[ \hat{\textnormal{y} } \right] - y \right)^2 + \mathbb{E} \left[ \hat{\textnormal{y} }^2 - 2 \mathbb{E} \left[ \hat{\textnormal{y} } \right]^2 + \mathbb{E} \left[ \hat{\textnormal{y} } \right]^2 \right] \\
    &= \left( \mathbb{E} \left[ \hat{\textnormal{y} } \right] - y \right)^2 + \mathbb{E} \left[ \hat{\textnormal{y} }^2 - 2 \hat{\textnormal{y} } \mathbb{E} \left[ \hat{\textnormal{y} } \right] + \mathbb{E} \left[ \hat{\textnormal{y} } \right]^2 \right] \\
    &= \left( \mathbb{E} \left[ \hat{\textnormal{y} } \right] - y \right)^2 + \mathbb{E} \left[ \left( \hat{\textnormal{y} } - \mathbb{E} \left[ \hat{\textnormal{y} } \right] \right)^2 \right] \\
    \mathbb{E} \left[ (y - \hat{\textnormal{y} })^2 \right]
    &= \underbrace{\left( \mathbb{E} \left[ \hat{\textnormal{y} } \right] - y \right)^2}_{\text{bias}^2} + \underbrace{\mathrm{Var} \left[ \hat{\textnormal{y} } \right]}_{\text{variance}}
\end{align*}

\section{Penalized models}
\label{sec:penalized}
Depending on the class of methods, there exist different strategies to tackle overfitting.

For neighbor methods, the number of neighbors used to define the neighborhood of any input and the strategy to compute the weights are the key hyperparameters to control the bias-variance trade-off.
For models that are presented in the remaining sections of this chapter, we mention strategies to address the bias-variance trade-off in their respective sections.
In this section, we present the most commonly used strategies for models whose parameters are optimized by minimizing a cost function defined as the mean loss values over all the training samples:
$$
\min_{\bm{w}} J(\bm{w}) \quad \text{with} \quad J(\bm{w}) = \frac{1}{n} {\sum_{i=1}^{n} \ell\left(y^{(i)}, f(\bm{x}^{(i)};\bm{w})\right)}
$$
This is for instance the case of the linear and logistic regression methods presented in the previous sections.

\subsection{Penalties}

The main idea is to introduce a \emph{penalty term} $\text{Pen}(\bm{w})$ that will constraint the parameters $\bm{w}$ to have some desired properties.
The most common penalties are the $\ell_2$ penalty, the $\ell_1$ penalty and the elastic-net penalty.

\subsubsection{\texorpdfstring{$\ell_2$}{} penalty}

The $\ell_2$ penalty is defined as the squared $\ell_2$ norm of the $\bm{w}$ coefficients:
$$
    \ell_2(\bm{w}) = \Vert \bm{w} \Vert_2^2 = \sum_{j=1}^p w_j^2 \\
$$
The $\ell_2$ penalty forces each coefficient $w_i$ not to be too large and makes the coefficients more robust to collinearity (i.e., when some features are approximately linear combinations of the other features).

\subsubsection{\texorpdfstring{$\ell_1$}{} penalty}

\begin{figure}
    \centering
    \includegraphics[width=0.6\textwidth]{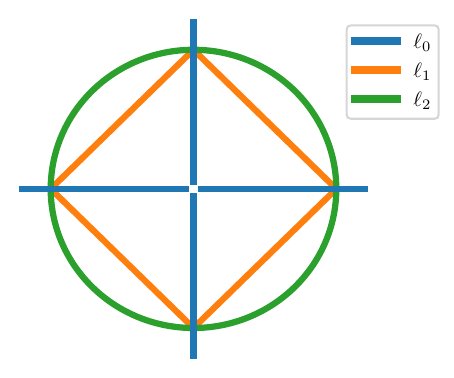}
    \caption{
    Unit balls of the $\ell_0$, $\ell_1$ and $\ell_2$ norms.
    For each norm, the set of points in $\mathbb{R}^2$ whose norm is equal to $1$ is plotted.
    The $\ell_1$ norm is the best convex approximation to the $\ell_0$ norm.
    Note that the lines for the $\ell_0$ norm extend to $-\infty$ and $+\infty$, but are cut for plotting reasons.
    }
    \label{fig:unit_balls}
\end{figure}

The $\ell_2$ penalty forces the values of the parameters not to be too large, but does not incentive to make small values tend to zero.
Indeed, the square of a small value is even smaller.
When the number of features is large, or when interpretability is important, it can be useful to make the model select the most important features.
The corresponding metric is the $\ell_0$ ``norm'' (which is not a proper norm in the mathematical sense), defined as the number of nonzero elements:
$$
\ell_0(\bm{w}) = \Vert \bm{w} \Vert_0 = \sum_{j=1}^p \bm{1}_{w_j \ne 0}
$$
However, the $\ell_0$ ``norm'' is neither differentiable nor convex (which are useful properties to solve an optimization problem, but this is not further detailed for the sake of conciseness).
The best convex differentiable approximation of the $\ell_0$ ``norm'' is the the $\ell_1$ norm (see \autoref{fig:unit_balls}), defined as the sum of the absolute values of each element:
$$
\ell_1(\bm{w}) = \Vert \bm{w} \Vert_1 = \sum_{j=1}^p \vert w_j \vert
$$

\subsubsection{Elastic-net penalty}

Both the $\ell_2$ and $\ell_1$ penalties have their upsides and downsides.
In order to try to obtain the best of penalties, one can add both penalties in the objective function.
The combination of both penalties is known as the \emph{elastic-net} penalty:
$$
    \text{EN}(\bm{w}, \alpha) = \alpha \Vert \bm{w} \Vert_1 + (1 - \alpha) \Vert \bm{w} \Vert_2^2
$$
where $\alpha \in [0, 1]$ is a hyperparameter representing the proportion of the $\ell_1$ penalty compared to the $\ell_2$ penalty.

\subsection{New optimization problem}

\begin{figure}
    \centering
    \includegraphics[width=1.\textwidth]{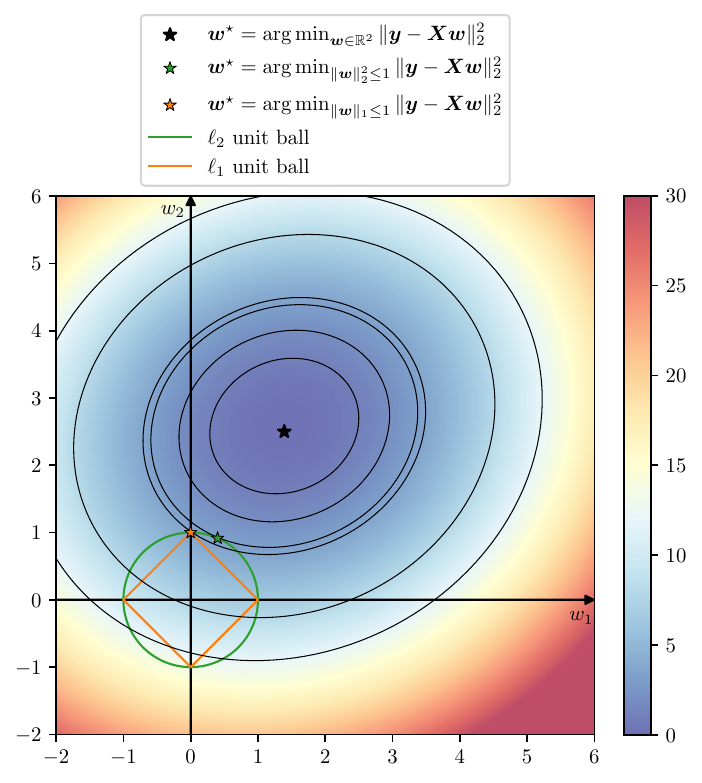}
    \caption{
    Illustration of the minimization problem with a constraint on the penalty term.
    The plot represents the value of the loss function for different values of the two coefficients for a linear regression task.
    The black star indicates the optimal solution with no constraint.
    The green and orange stars indicate the optimal solutions when imposing a constraint on the $\ell_2$ and $\ell_1$ norms of the parameters $\bm{w}$ respectively.
    }
    \label{fig:regularization_constraint}
\end{figure}

A natural approach would be to add a constraint to the minimization problem:
\begin{equation}
    \label{eq:minimization_constraint}
    \min_{\bm{w}} J(\bm{w}) \qquad \text{subject to} \qquad \text{Pen}(\bm{w}) < c
\end{equation}
which reads as ``Find the optimal parameters that minimize the cost function $J$ among all the parameters $\bm{w}$ that satisfy $\text{Pen}(\bm{w}) < c$'' for a positive real number $c$.
\autoref{fig:regularization_constraint} illustrates the optimal solution of a simple linear regression task with different constraints.
This figure also highlights the sparsity property of the $\ell_1$ penalty (the optimal parameter for the horizontal axis is set to zero) that the $\ell_2$ penalty does not have (the optimal parameter for the horizontal axis is small but different from zero).

Although this approach is appealing due to its intuitiveness and the possibility to set the maximum possible penalty on the parameters $\bm{w}$, it leads to a minimization problem that is not trivial to solve.
A similar approach consists in adding the regularization term in the cost function:
\begin{equation}
    \label{eq:minimization_penalty}
    \min_{\bm{w}} J(\bm{w}) + \lambda \times \text{Pen}(\bm{w})
\end{equation}
where $\lambda > 0$ is a hyperparameter that controls the weights of the penalty term compared to the mean loss values over all the training samples.
This formulation is related to the Lagrangian function of the minimization problem with the penalty constraint.

This formulation leads to a minimization problem with no constraint which is much easier to solve.
One can actually show that \autoref{eq:minimization_constraint} and \autoref{eq:minimization_penalty} are related: solving \autoref{eq:minimization_penalty} for a given $\lambda$, whose optimal solution is denoted by $\bm{w}_{\lambda}^{\star}$, is equivalent to solving \autoref{eq:minimization_constraint} for $c = \text{Pen}(\bm{w}_{\lambda}^{\star})$.
In other words, solving \autoref{eq:minimization_penalty} for a given $\lambda$ is equivalent to solving \autoref{eq:minimization_constraint} for $c$ whose value is only known after finding the optimal solution of \autoref{eq:minimization_penalty}.

\begin{figure}[hbtp]
    \centering
    \includegraphics[width=\textwidth]{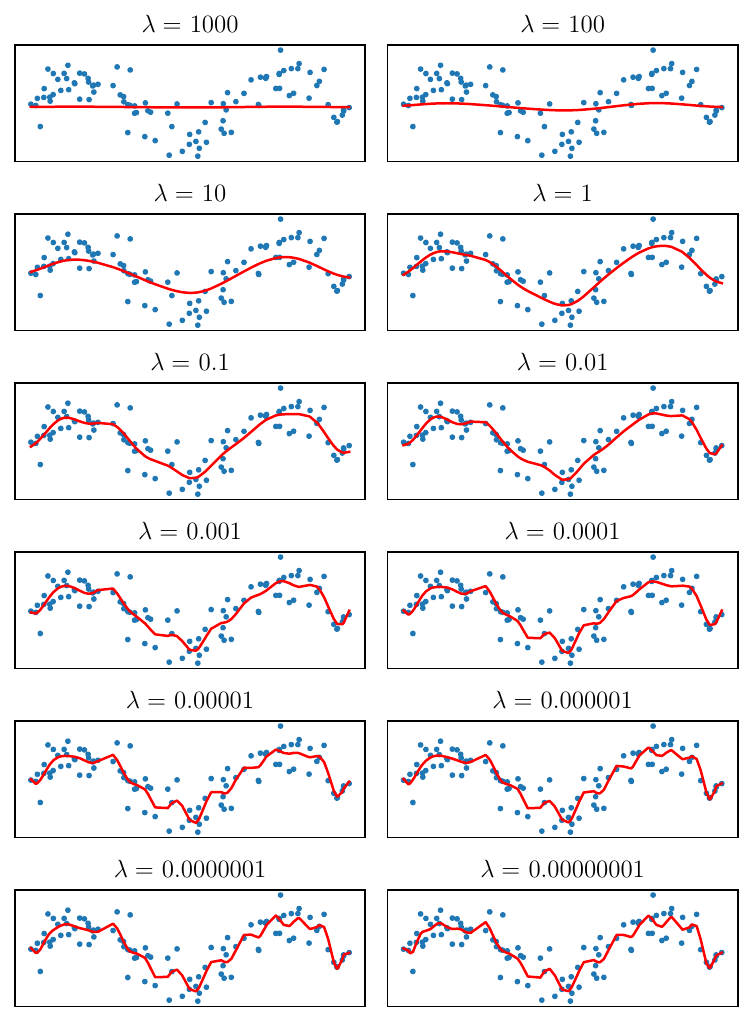}
    \caption{
    Illustration of regularization.
    A kernel ridge regression algorithm is fitted on the training data (blue points) with different values of $\lambda$, which is the weight of the regularization in the cost function.
    The smaller values of $\lambda$, the smaller the weight of the $\ell_2$ regularization.
    The algorithm underfits (respectively overfits) the data when the value of $\lambda$ is too large (respectively low).
    }
    \label{fig:regularization_kernel_ridge_regression}
\end{figure}
\autoref{fig:regularization_kernel_ridge_regression} illustrates the impact of the regularization term $\lambda \times \text{Pen}(\bm{w})$ on the prediction function of a kernel ridge regression algorithm (see \autoref{sec:kernel_methods} for more details) for different values of $\lambda$.
For high values of $\lambda$, the regularization term is dominating the mean loss value, making the prediction function not fitting well enough the training data (underfitting).
For small values of $\lambda$, the mean loss value is dominating the regularization term, making the prediction function fitting too well the training data (overfitting).
A good balance between the mean loss value and the regularization term is required to learn the best function.

Since linear regression is one of the oldest and best-known models, the aforementioned penalties were originally introduced for linear regression:
\begin{itemize}
    \item Linear regression with the $\ell_2$ penalty is also known as ridge \citep{tikhonov_solutions_1977}: 
    $$
    \min_{\bm{w}} \Vert \bm{y} - \bm{X} \bm{w} \Vert_2^2 + \lambda \Vert \bm{w} \Vert_2^2
    $$
    As in ordinary least squares regression, there exists a closed formula for the optimal solution:
    $$
    \bm{w}^\star = \left(\bm{X}^\top \bm{X} + \lambda \bm{I} \right)^{-1} \bm{X}^\top \bm{y}
    $$
    \item Linear regression with the $\ell_1$ penalty is also known as lasso \cite{tibshirani_regression_1996}:
    $$
    \min_{\bm{w}} \Vert \bm{y} - \bm{X} \bm{w} \Vert_2^2 + \lambda \Vert \bm{w} \Vert_1
    $$
    \item Linear regression with the elastic-net penalty is also known as elastic-net \cite{zou_regularization_2005}:
    $$
    \min_{\bm{w}} \Vert \bm{y} - \bm{X} \bm{w} \Vert_2^2 + \lambda \alpha \Vert \bm{w} \Vert_1 + \lambda (1 - \alpha) \Vert \bm{w} \Vert_2^2
    $$
\end{itemize}
The penalties can also be added in other models such as logistic regression, support-vector machines, artificial neural networks, etc.

\section{Support-vector machine}
\label{sec:support_vector_machine}

Linear and logistic regression take into account every training sample in order to find the best line, which is due to their corresponding loss functions: the squared error is zero only if the true and predicted outputs are equal, and the logistic loss is always positive.
One could argue that the training samples whose outputs are ``easily'' well predicted are not relevant: only the training samples whose outputs are not ``easily'' well predicted or are wrongly predicted should be taken into account.
The support vector machine (SVM) is based on this principle.

\subsection{Original formulation}

\begin{figure}
\centering
    \includegraphics[width=1.\textwidth]{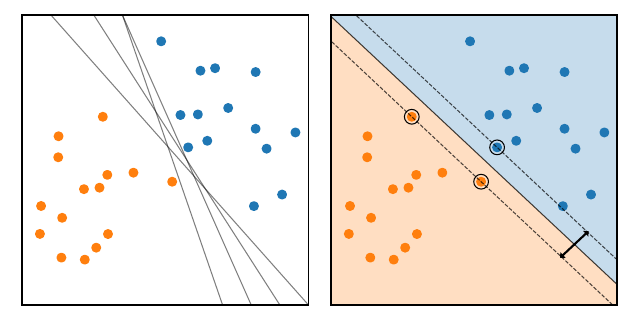}
    \caption{Support vector machine classifier with linearly separable classes.
    When two classes are linearly separable, there exists an infinite number of hyperplanes separating them (left).
    The decision function of the support vector machine classifier is the hyperplane that maximizes the margin, that is the distance between the hyperplane and the closest points to the hyperplane (right).
    Support vectors are highlighted with a black circle surrounding them.}
\label{fig:classification_svm_linearly_separable}
\end{figure}

The original support vector machine was invented in 1963 and was a linear binary classification method \citep{vapnik_pattern_1963}.
\autoref{fig:classification_svm_linearly_separable} illustrates the main concept of its original version.
When both classes are linearly separable, there exists an infinite number of hyperplanes that separate both classes.
The SVM finds the hyperplane that maximizes the \emph{margin}, that is the distance between the hyperplane and the closest points of both classes to the hyperplane, while linearly separating both classes.

\begin{figure}
    \centering
    \includegraphics[width=1.\textwidth]{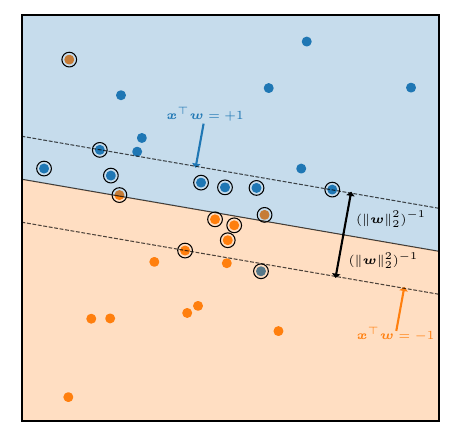}
    \caption{Decision function of a support-vector machine classifier with a linear kernel when both classes are not strictly linearly separable. The support vectors are the training points within the margin of the decision function and the misclassified training points. The support vectors are highlighted with a black circle surrounding them.}
    \label{fig:svm_classification_not_linearly_separable}
\end{figure}

The SVM was later updated to non-separable classes \citep{cortes_support-vector_1995}.
\autoref{fig:svm_classification_not_linearly_separable} illustrates the role of the margin in this case.
The dashed lines correspond to the hyperplanes defined by the equations $\bm{x}^\top \bm{w} = +1$ and $\bm{x}^\top \bm{w} = -1$.
The margin is the distance between both hyperplanes and is equal to $2 / \Vert \bm{w} \Vert_2^2 $.
It defines which samples are included in the decision function of the model: a sample is included if and only if it is inside the margin, or outside the margin and misclassified.
Such samples are called \textit{support vectors} and are illustrated in \autoref{fig:svm_classification_not_linearly_separable} with a black circle surrounding them.
In this case, the margin can be seen a regularization term: the larger the margin is, the more support vectors are included in the decision function, the more regularized the model is.

\begin{figure}
    \centering
    \includegraphics[width=1.\textwidth]{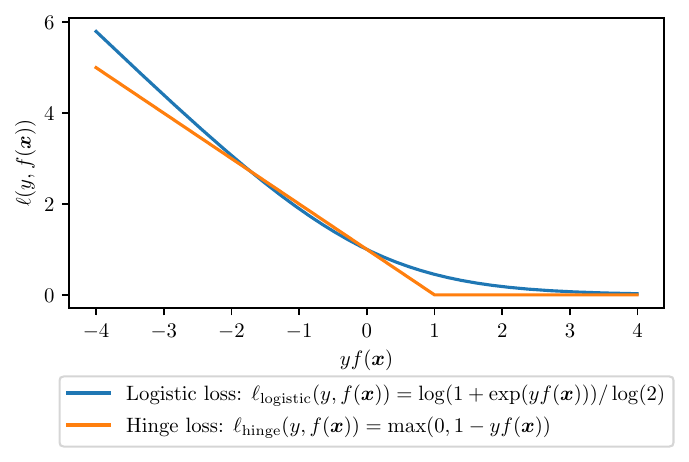}
    \caption{
        Binary classification losses.
        The logistic loss is always positive, even when the point is accurately classified with high confidence (i.e., when $y f(\bm{x}) \gg 0$), whereas the hinge loss is equal to zero when the point is accurately classified with good confidence (i.e., when $y f(\bm{x}) \geq 1$).
        }
    \label{fig:classification_losses}
\end{figure}

The loss function for the SVM is called the \emph{hinge loss} and is defined as:
$$
\ell_{\text{hinge}}(y, f(\bm{x})) = \max(0, 1 - y f(\bm{x}))
$$
\autoref{fig:classification_losses} illustrates the curves of the logistic and hinge losses.
The logistic loss is always positive, even when the point is accurately classified with high confidence (i.e., when $y f(\bm{x}) \gg 0$), whereas the hinge loss is equal to zero when the point is accurately classified with good confidence (i.e., when $y f(\bm{x}) \geq 1$).
One can see that a sample $(\bm{x}, y)$ is a support vector if and only if $y f(\bm{x}) \geq = 1$, that is if and only if $\ell_{\text{hinge}}(y, f(\bm{x})) = 0$.

The optimal $\bm{w}$ coefficients for the original version are estimated by minimizing an objective function consisting of the sum of the \emph{hinge loss} values and a $\ell_2$ penalty term (which is inversely proportional to the margin):
$$
\min_{\bm{w}} \sum_{i=1}^n \max(0, 1 - y^{(i)} \bm{x}^{(i)\top} \bm{w}) + \frac{1}{2C} \Vert \bm{w} \Vert_2^2
$$

\subsection{General formulation with kernels}

The SVM was later updated to non-linear decision functions with the use of \emph{kernels} \citep{boser_training_1992}.

In order to have a non-linear decision function, one could map the input space $\mathcal{I}$ into another space (often called the \emph{feature space}), denoted by $\mathcal{G}$, using a function denoted by $\phi$:
\begin{align*}
  \phi \colon \mathcal{I} &\to \mathcal{G} \\
  \bm{x} &\mapsto \phi(\bm{x})
\end{align*}
The decision function would still be linear (with a dot product), but in the feature space:
$$
f(\bm{x}) = \phi(\bm{x})^\top \bm{w}
$$
Unfortunately, solving the corresponding minimization problem is not trivial:
\begin{equation}
    \label{eq:svm_kernel}
    \min_{\bm{w}} \sum_{i=1}^n \max \left(0, 1 - y^{(i)} \phi(\bm{x}^{(i)})^\top \bm{w} \right) + \frac{1}{2C} \Vert \bm{w} \Vert_2^2
\end{equation}
Nonetheless, two mathematical properties make the use of non-linear transformations in the feature space possible: the \emph{kernel trick} and the \emph{representer theorem}.

The kernel trick asserts that the dot product in the feature space can be computed using only the points from the input space and a \emph{kernel function}, denoted by $K$:
$$
\forall \bm{x}, \bm{x}' \in \mathcal{I},\, \phi(\bm{x})^\top \phi(\bm{x}') = K(\bm{x}, \bm{x}')
$$

The representer theorem \cite{aizerman_theoretical_1967, scholkopf_generalized_2001} asserts that, under certain conditions on the kernel $K$ and the feature space $\mathcal{G}$ associated with the function $\phi$, any minimizer of \autoref{eq:svm_kernel} admits the following form:
$$
f = \sum_{i=1}^n \alpha_i K(\cdot, \bm{x}^{(i)})
$$
where $\bm{\alpha}$ solves:
$$
\min_{\bm{\alpha}} \sum_{i=1}^n \max(0, 1 - y^{(i)} [\bm{K} \bm{\alpha}]_i) + \frac{1}{2C} \bm{\alpha}^\top \bm{K} \bm{\alpha}
$$
where $\bm{K}$ is the $n \times n$ matrix consisting of the evaluations of the kernel on all the pairs of training samples: $\forall i, j \in \{1, \ldots, n \}, \, K_{ij} = K(\bm{x}^{(i)}, \bm{x}^{(j)})$.

Because the hinge loss is equal to zero if and only if $y f(\bm{x})$ is greater than or equal to $1$, only the training samples $(\bm{x}^{(i)}, y^{(i)})$ such that $y^{(i)} f(\bm{x}^{(i)}) < 1$ have a nonzero $\alpha_i$ coefficient.
These points are the so-called support vectors and this is why they are the only training samples contributing to the decision function of the model:
\begin{gather*}
    \text{SV} = \{ i \in \{1, \ldots, n\}  \,|\, \alpha_i \neq 0 \} \\
    f(\bm{x}) = \sum_{i=1}^n \alpha_i K(\bm{x}, \bm{x}^{(i)}) = \sum_{i \in \text{SV}} \alpha_i K(\bm{x}, \bm{x}^{(i)})
\end{gather*}

\begin{floatbox}[t]
    \begin{nicebox}[Support-vector machine]
        \label{box:support_vector_machine}
        \begin{itemize}[leftmargin=2mm]
            \item \textbf{Main idea}: hyperplane (i.e., line) that maximizes the margin (i.e., the distance between the hyperplane and the closest inputs to the hyperplane).
            \item \textbf{Support vectors}: only the misclassified inputs and the inputs well classified but with low confidence are taken into account.
            \item \textbf{Non-linearity}: decision function can be non-linear with the use of non-linear kernels.
            \item \textbf{Regularization}: $\ell_2$ penalty.
        \end{itemize}
    \end{nicebox}
\end{floatbox}

\begin{figure}
    \centering
    \includegraphics[width=1.\textwidth]{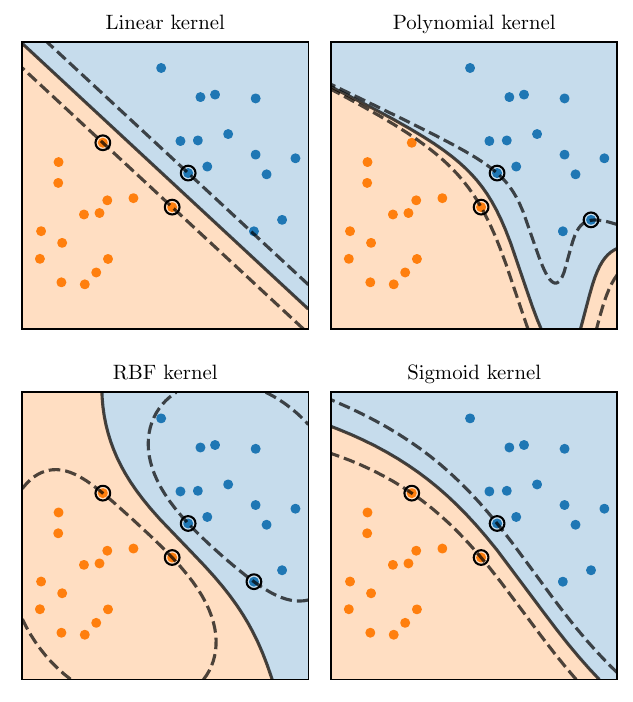}
    \caption{Impact of the kernel on the decision function of a support vector machine classifier. A non-linear kernel allows for a non-linear decision function.}
    \label{fig:classification_svm_kernels}
\end{figure}

The kernel trick and the representer theorem show that it is more practical to work with the kernel $K$ instead of the mapping function $\phi$.
Popular kernel functions include:
\begin{itemize}
    \item the linear kernel:
        $$K(\bm{x}, \bm{x}') = \bm{x}^\top \bm{x}' $$
    \item the polynomial kernel:
        $$K(\bm{x}, \bm{x}') = \left( \gamma \, \bm{x}^\top \bm{x}' + c_0 \right)^d \quad \text{with} \quad \gamma > 0, \, c_0 \ge 0, \, d \in \mathbb{N}^* $$
    \item the sigmoid kernel:
        $$K(\bm{x}, \bm{x}') = \tanh \left( \gamma \, \bm{x}^\top \bm{x}' + c_0 \right) \quad \text{with} \quad \gamma > 0, \, c_0 \ge 0$$
    \item the radial basis function (RBF) kernel:
        $$K(\bm{x}, \bm{x}') = \exp \left( - \gamma \, \Vert \bm{x} - \bm{x}' \Vert_2^2 \right) \quad \text{with} \quad \gamma > 0$$
\end{itemize}
The linear kernel yields a linear decision function and is actually identical to the original formulation of the SVM (one can show that there is a mapping between the $\bm{\alpha}$ and $\bm{w}$ coefficients).
Non-linear kernels allow for non-linear, more complex, decision functions.
This is particularly useful when the data is not linearly separable, which is the most common use case.
\autoref{fig:classification_svm_kernels} illustrates the decision function and the margin of a SVM classification model for four different kernels.

The SVM was also extended to regression tasks with the use of the \emph{$\varepsilon$-insensitive loss}.
Similarly to the hinge loss, which is equal to zero for points that are correctly classified and outside the margin, the $\varepsilon$-insensitive loss is equal to zero when the error between the true target value and the predicted value is not greater than $\varepsilon$:
$$
\ell_{\varepsilon-\text{insensitive}}(y, f(\bm{x})) = \max(0, \vert y - f(\bm{x}) \vert - \varepsilon)
$$
The objective function for the SVM regression method combines the values of $\varepsilon$-insensitive loss of the training points and the $\ell_2$ penalty:
$$
\min_{\bm{w}} \sum_{i=1}^n \max \left(0, \left\vert y^{(i)} - \phi(\bm{x}^{(i)})^\top \bm{w} \right\vert - \varepsilon \right) + \frac{1}{2C} \Vert \bm{w} \Vert_2^2
$$

\begin{figure}
    \centering
    \includegraphics[width=1.\textwidth]{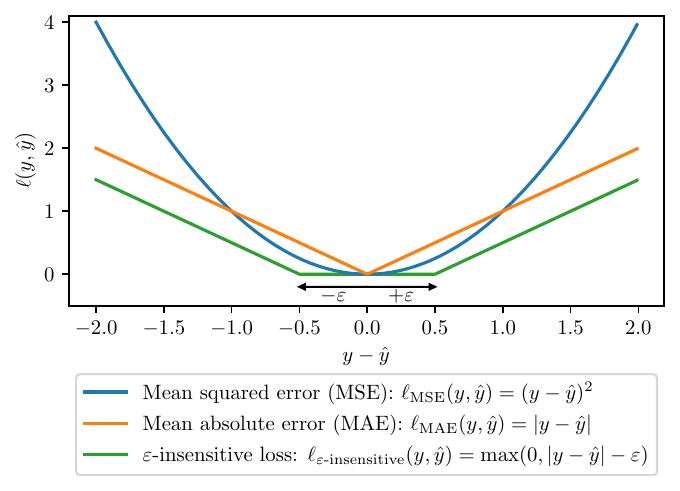}
    \caption{Regression losses. The squared error loss takes very small values for small errors and very large values for large errors, whereas the absolute error loss takes small values for small errors and large values for large errors.
    Both losses take small but non-zero values when the error is small.
    On the contrary, the $\varepsilon$-insensitive loss is null when the error is small and otherwise equal the absolute error loss minus $\varepsilon$. When computed over several samples, the squared and absolute error losses are often referred to as mean squared error (MSE) and mean absolute error (MAE) respectively.
    }
    \label{fig:regression_losses}
\end{figure}

\autoref{fig:regression_losses} illustrates the curves of three regression losses.
The squared error loss takes very small values for small errors and very high values for high errors, whereas the absolute error loss takes small values for small errors and high values for high errors. Both losses take small but non-zero values when the error is small.
On the contrary, the $\varepsilon$-insensitive loss is null when the error is small and otherwise equal to the absolute error loss minus $\varepsilon$.

\section{Multiclass classification}
\label{sec:multiclass}
The classification methods that we presented so far, logistic regression and support-vector machines, are binary classifiers: they can only be used when there are only two possible outcomes.
However, in practice, it is common to have more than two possible outcomes.
For instance, differential diagnosis of brain disorders is often between several, and not only two, diseases.

\begin{figure}
    \centering
    \begin{tikzpicture}[
    empty/.style = {rectangle, font=\footnotesize},
    block/.style = {draw, rectangle, font=\footnotesize},
    block head/.style={block, font=\normalsize, draw=C3!80, fill=C3!5, very thick},
    block positive/.style = {block, fill=C0!20},
    block negative/.style = {block, fill=C1!20},
]

\node[block head, from={-6.4,0 to -2.4,1}] {One-vs-rest};

\node[block positive, from={-6.4,-1 to -5.4,-0.5}] {$\{1\}$};
\node[empty, from={-5.4,-1 to -4.4,-0.5}]{vs.};
\node[block negative, from={-4.4,-1 to -2.4,-0.5}] {$\{2, 3, 4, 5\}$};

\node[block positive, from={-6.4,-1.8 to -5.4,-1.3}] {$\{2\}$};
\node[empty, from={-5.4,-1.8 to -4.4,-1.3}]{vs.};
\node[block negative, from={-4.4,-1.8 to -2.4,-1.3}] {$\{1, 3, 4, 5\}$};

\node[block positive, from={-6.4,-2.6 to -5.4,-2.1}] {$\{3\}$};
\node[empty, from={-5.4,-2.6 to -4.4,-2.1}]{vs.};
\node[block negative, from={-4.4,-2.6 to -2.4,-2.1}] {$\{1, 2, 4, 5\}$};

\node[block positive, from={-6.4,-3.4 to -5.4,-2.9}] {$\{4\}$};
\node[empty, from={-5.4,-3.4 to -4.4,-2.9}]{vs.};
\node[block negative, from={-4.4,-3.4 to -2.4,-2.9}] {$\{1, 2, 3, 5\}$};

\node[block positive, from={-6.4,-4.2 to -5.4,-3.7}] {$\{5\}$};
\node[empty, from={-5.4,-4.2 to -4.4,-3.7}]{vs.};
\node[block negative, from={-4.4,-4.2 to -2.4,-3.7}] {$\{1, 2, 3, 4\}$};

\node[block head, from={-2,0 to 2,1}] {One-vs-one};

\node[block positive, from={-1.5,-1.0 to -0.5,-0.5}] {$\{1\}$};
\node[empty, from={-0.5,-1.0 to 0.5,-0.5}]{vs.};
\node[block negative, from={0.5,-1.0 to 1.5,-0.5}] {$\{2\}$};

\node[block positive, from={-1.5,-1.8 to -0.5,-1.3}] {$\{1\}$};
\node[empty, from={-0.5,-1.8 to 0.5,-1.3}]{vs.};
\node[block negative, from={0.5,-1.8 to 1.5,-1.3}] {$\{3\}$};

\node[block positive, from={-1.5,-2.6 to -0.5,-2.1}] {$\{1\}$};
\node[empty, from={-0.5,-2.6 to 0.5,-2.1}]{vs.};
\node[block negative, from={0.5,-2.6 to 1.5,-2.1}] {$\{4\}$};

\node[block positive, from={-1.5,-3.4 to -0.5,-2.9}] {$\{1\}$};
\node[empty, from={-0.5,-3.4 to 0.5,-2.9}]{vs.};
\node[block negative, from={0.5,-3.4 to 1.5,-2.9}] {$\{5\}$};

\node[block positive, from={-1.5,-4.2 to -0.5,-3.7}] {$\{2\}$};
\node[empty, from={-0.5,-4.2 to 0.5,-3.7}]{vs.};
\node[block negative, from={0.5,-4.2 to 1.5,-3.7}] {$\{3\}$};

\node[block positive, from={-1.5,-5.0 to -0.5,-4.5}] {$\{2\}$};
\node[empty, from={-0.5,-5.0 to 0.5,-4.5}]{vs.};
\node[block negative, from={0.5,-5.0 to 1.5,-4.5}] {$\{4\}$};

\node[block positive, from={-1.5,-5.8 to -0.5,-5.3}] {$\{2\}$};
\node[empty, from={-0.5,-5.8 to 0.5,-5.3}]{vs.};
\node[block negative, from={0.5,-5.8 to 1.5,-5.3}] {$\{5\}$};

\node[block positive, from={-1.5,-6.6 to -0.5,-6.1}] {$\{3\}$};
\node[empty, from={-0.5,-6.6 to 0.5,-6.1}]{vs.};
\node[block negative, from={0.5,-6.6 to 1.5,-6.1}] {$\{4\}$};

\node[block positive, from={-1.5,-7.4 to -0.5,-6.9}] {$\{3\}$};
\node[empty, from={-0.5,-7.4 to 0.5,-6.9}]{vs.};
\node[block negative, from={0.5,-7.4 to 1.5,-6.9}] {$\{5\}$};

\node[block positive, from={-1.5,-8.2 to -0.5,-7.7}] {$\{4\}$};
\node[empty, from={-0.5,-8.2 to 0.5,-7.7}]{vs.};
\node[block negative, from={0.5,-8.2 to 1.5,-7.7}] {$\{5\}$};

\node[block head, from={2.4,0 to 6.4,1}] {Output code};

\node[block positive, from={2.4,-1.0 to 3.7,-0.5}] {$\{1, 3\}$};
\node[empty, from={3.7,-1.0 to 4.7,-0.5}]{vs.};
\node[block negative, from={4.7,-1.0 to 6.4,-0.5}] {$\{2, 4, 5\}$};

\node[block positive, from={2.4,-1.8 to 4.1,-1.3}] {$\{1, 4, 5\}$};
\node[empty, from={4.1,-1.8 to 5.1,-1.3}]{vs.};
\node[block negative, from={5.1,-1.8 to 6.4,-1.3}] {$\{2, 3\}$};

\node[block positive, from={2.6,-2.6 to 3.4,-2.1}] {$\{2\}$};
\node[empty, from={3.4,-2.6 to 4.4,-2.1}]{vs.};
\node[block negative, from={4.4,-2.6 to 6.4,-2.1}] {$\{1, 3, 4, 5\}$};

\node[block positive, from={2.4,-3.4 to 4.1,-2.9}] {$\{1, 2, 3\}$};
\node[empty, from={4.1,-3.4 to 5.1,-2.9}]{vs.};
\node[block negative, from={5.1,-3.4 to 6.4,-2.9}] {$\{4, 5\}$};

\node[block positive, from={2.4,-4.2 to 3.7,-3.7}] {$\{2, 5\}$};
\node[empty, from={3.7,-4.2 to 4.7,-3.7}]{vs.};
\node[block negative, from={4.7,-4.2 to 6.4,-3.7}] {$\{1, 3, 4\}$};

\node[block positive, from={2.4,-5.0 to 4.1,-4.5}] {$\{2, 3, 4\}$};
\node[empty, from={4.1,-5.0 to 5.1,-4.5}]{vs.};
\node[block negative, from={5.1,-5.0 to 6.4,-4.5}] {$\{1, 5\}$};

\node[block positive, from={2.4,-5.8 to 3.4,-5.3}] {$\{4\}$};
\node[empty, from={3.4,-5.8 to 4.4,-5.3}]{vs.};
\node[block negative, from={4.4,-5.8 to 6.4,-5.3}] {$\{1, 2, 3, 5\}$};

\node[empty, from={2.4,-6.6 to 3.9,-6.1}] {$\vdots$};
\node[empty, from={3.9,-6.6 to 4.9,-6.1}]{$\vdots$};
\node[empty, from={4.9,-6.6 to 6.4,-6.1}] {$\vdots$};

\draw[black, thick, dashed] (-2.2, 1) -- (-2.2,-8.5);
\draw[black, thick, dashed] (2.2, 1) -- (2.2,-8.5);

\end{tikzpicture}
    \caption{
    Main approaches to convert a multiclass classification task into several binary classification tasks.
    In the one-vs-rest approach, each class is associated to a binary classification model that is trained to separate this class from all the other classes.
    In the one-vs-one approach, a binary classifier is trained on each pair of classes.
    In the error correcting output code approach, the classes are (randomly) split into two groups and a binary classifier is trained for each split.
    }
\label{fig:multiclass_strategies}
\end{figure}
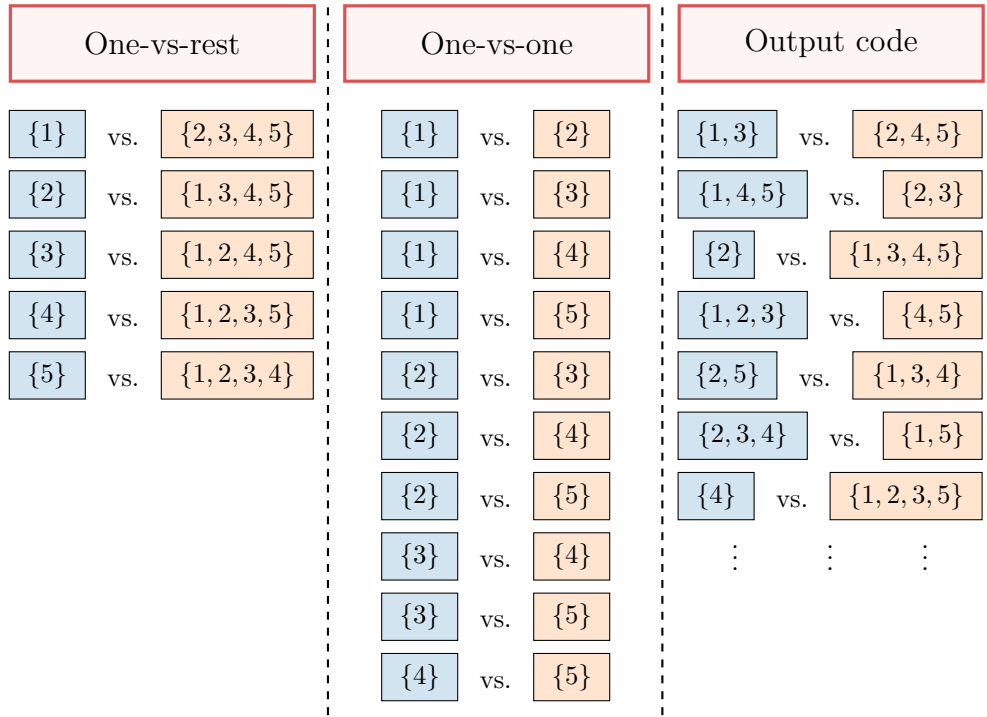

Several strategies have been proposed to extend any binary classification method to multiclass classification tasks.
They all rely on transforming the multiclass classification task into several binary classification tasks.
In this section, we present the most commonly used strategies: \emph{one-vs-rest}, \emph{one-vs-one} and \emph{error correcting output code} \cite{aly_survey_2005}.
\autoref{fig:multiclass_strategies} illustrates the main ideas of these approaches.
But first, we present a natural extension of logistic regression to multiclass classification tasks which is often referred to as \emph{multinomial logistic regression} \cite{bishop_pattern_2006}.

\subsection{Multinomial logistic regression}

For binary classification, logistic regression is characterized by a hyperplane: the signed distance to the hyperplane is mapped into the probability of belonging to the positive class using the sigmoid function.
However, for multiclass classification, a single hyperplane is not enough to characterize all the classes.
Instead, each class $\mathcal{C}_k$ is characterized by a hyperplane $\bm{w}_k$ and, for any input $\bm{x}$, one can compute the signed distance $\bm{x}^\top \bm{w}_k$ between the input $\bm{x}$ and the hyperplane $\bm{w}_k$.
The signed distances are mapped into probabilities using the softmax function, defined as $\textnormal{softmax} \left(x_1, \ldots, x_q \right)= \left( \frac{ \exp\left( x_1 \right) }{ \sum_{j=1}^q \exp\left( x_j \right) }, \ldots, \frac{ \exp\left( x_q \right) }{ \sum_{j=1}^q \exp\left( x_j \right) } \right)$, as follows:
$$
\forall k \in \{1, \ldots, q\}, \, P (\textnormal{y} = \mathcal{C}_k | \mathbf{x} = \bm{x} ) =\frac{ \exp\left( \bm{x}^\top \bm{w}_k \right) }{ \sum_{j=1}^q \exp\left( \bm{x}^\top \bm{w}_j \right) }
$$

The coefficients $(\bm{w}_k)_{1 \leq k \leq q}$ are still estimated by maximizing the likelihood function:
$$
L(\bm{w}_1, \ldots, \bm{w}_q) = \prod_{i=1}^n \prod_{k=1}^q P \left( \textnormal{y} = \mathcal{C}_k | \mathbf{x} = \bm{x}^{(i)} \right)^{\bm{1}_{y^{(i)} = \mathcal{C}_k}}
$$
which is equivalent to minimizing the negative log-likelihood:
\begin{align*}
    &- \log( L(\bm{w}_1, \ldots, \bm{w}_q) ) \\
    &= - \sum_{i=1}^n \sum_{k=1}^q \bm{1}_{y^{(i)} = \mathcal{C}_k} \log \left( P \left( \textnormal{y} = \mathcal{C}_k | \mathbf{x} = \bm{x}^{(i)} \right) \right) \\
    &= \sum_{i=1}^n - \sum_{k=1}^q \bm{1}_{y^{(i)} = \mathcal{C}_k} \log \left( \frac{ \exp\left( \bm{x^{(i)\top}} \bm{w}_k \right) }{ \sum_{j=1}^q \exp\left( \bm{x^{(i)\top}} \bm{w}_j \right) } \right) \\
    &= \sum_{i=1}^n \ell_{\textnormal{cross-entropy}} \left( y^{(i)},  \textnormal{softmax} \left(\bm{x^{(i)\top}} \bm{w}_1, \ldots, \bm{x^{(i)\top}} \bm{w}_q \right) \right)
\end{align*}
where $\ell_{\textnormal{cross entropy}}$ is known as the \emph{cross-entropy loss} and is defined, for any label $y$ and any vector of probabilities $(\pi_1, \ldots, \pi_q)$, as:
$$
\ell_{\textnormal{cross-entropy}}( y, (\pi_1, \ldots, \pi_q) ) = - \sum_{k=1}^q \bm{1}_{y = \mathcal{C}_k} \pi_k
$$
This loss is commonly used to train artificial neural networks on classification tasks and is equivalent to the logistic loss in the binary case.

\begin{figure}
    \centering
    \includegraphics[width=\textwidth]{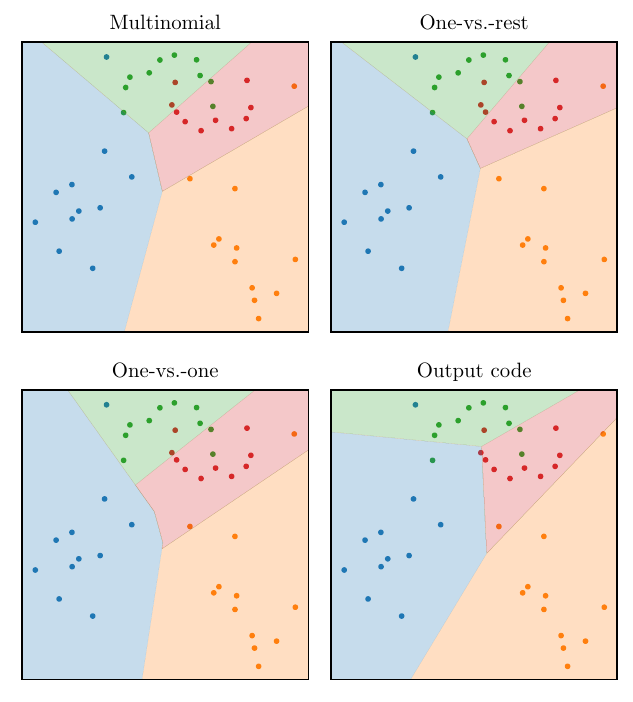}
    \caption{Illustration of the impact of the strategy used to handle a multiclass classification task on the decision function of a logistic regression model.}
    \label{fig:multiclass_logistic_regression}
\end{figure}

\autoref{fig:multiclass_logistic_regression} illustrates the impact of the strategy used to handle a multiclass classification task on the decision function.

\subsection{One-vs-rest}

A strategy to transform a multiclass classification task into several binary classification tasks is to fit a binary classifier for each class: the positive class is the given class, and the negative class consists of all the other classes merged into a single class.
This strategy is known as \emph{one-vs-rest}.
The advantage of this strategy is that each class is characterized by a single model, so that it is possible to gain deeper knowledge about the class by inspecting its corresponding model.
A consequence is that the predictions for new samples take into account the confidence of the models: the predicted class for a new input is the class for which the corresponding model is the most confident that this input belongs to its class.
The one-vs-rest strategy is the most commonly used strategy and usually a good default choice.

\subsection{One-vs-one}

Another strategy is to fit a binary classifier for each pair of classes: this strategy is known as \emph{one-vs-one}.
The advantage of this strategy is that the classes in each binary classification task are ``pure'', in the sense that different classes are never merged into a single class.
However, the number of binary classifiers that needs to be trained is larger for the one-vs-one strategy ($\frac{1}{2} q (q - 1)$) than for the one-vs-rest strategy ($q$).
Nonetheless, for the one-vs-one strategy, the number of training samples in each binary classification tasks is smaller than the total number of samples, which makes training each binary classifier usually faster.
Another drawback is that this strategy is less interpretable compared to the one-vs-rest strategy, as the predicted class corresponds to the class obtaining the most votes (i.e., winning the most one-vs-one matchups), which does not take into account the confidence in winning each matchup.\footnote{The confidences are actually taken into account but only in the event of a tie.}
For instance, winning a one-vs-one matchup with 0.99 probability gives the same result as winning the same matchup with 0.51 probability, i.e. one vote.

\subsection{Error correcting output code}

A substantially different strategy, inspired by the theory of error correction code, consists in merging a subset of classes into one class and the other subset into the other class, for each binary classification task.
This data is often called the code book and can be represented as a matrix whose rows correspond to the classes and whose columns correspond to the binary classification tasks.
The matrix consists only of $-1$ and $+1$ values that represent the corresponding label for each class and for each binary task.\footnote{The values are $0$ and $1$ when the classifier does not return scores but only probabilities.}
For any input, each binary classifier returns the score (or probability) associated with the positive class.
The predicted class for this input is the class whose corresponding vector is the most similar to the vector of scores, with similarity being assessed with the Euclidean distance (the lower, the more similar).
There exist advanced strategies to define the code book, but it has been argued than a random code book usually gives as good results as a sophisticated one \cite{james_error_1998}.

\section{Decision functions with normal distributions}
\label{sec:decision_functions_normal_distributions}

Normal distributions are popular distributions because they are commonly found in nature.
For instance, the distribution of heights and birth weights of human beings can be approximated using normal distributions.
Moreover, normal distributions are particularly easy to work with from a mathematical point of view.
For these reasons, a common model consists in assuming that the training input vectors are independently sampled from normal distributions.

A possible classification model consists in assuming that, for each class, all the corresponding inputs are sampled from a normal distribution with mean vector $\bm{\mu}_k$ and covariance matrix $\bm{\Sigma}_k$:
$$
\forall i \textnormal{ such that } y^{(i)} = \mathcal{C}_k,\, \bm{x}^{(i)} \sim \mathcal{N}(\bm{\mu}_k, \bm{\Sigma}_k)
$$
Using the probability density function of a normal distribution, one can compute the probability density of any input $\bm{x}$ associated to the distribution $\mathcal{N}(\bm{\mu}_k, \bm{\Sigma}_k)$ of class $\mathcal{C}_k$:
$$
p_{\mathbf{x} | \textnormal{y} = \mathcal{C}_k}(\bm{x}) = \frac{1}{\sqrt{(2\pi)^p |\bm{\Sigma}_k|}} \exp\left( - \frac{1}{2} [\bm{x} - \bm{\mu}_k]^\top \bm{\Sigma}_k^{-1} [\bm{x} - \bm{\mu}_k] \right)
$$

With such a probabilistic model, it is easy to compute the probability that a sample belongs to class $\mathcal{C}_k$ using Bayes rule:
$$
P(\textnormal{y} = \mathcal{C}_k | \mathbf{x} = \bm{x})
= \frac{ p_{\mathbf{x} | \textnormal{y} = \mathcal{C}_k}(\bm{x}) P(\textnormal{y} = \mathcal{C}_k) }{ p_{\mathbf{x}}(\bm{x}) }
$$
With normal distributions, it is mathematically easier to work with log-probabilities:
\begin{align}
\begin{split}
    &\log P(\textnormal{y} = \mathcal{C}_k | \mathbf{x} = \bm{x}) \\
    &= \log p_{\mathbf{x} | \textnormal{y} = \mathcal{C}_k}(\bm{x}) + \log P(\textnormal{y} = \mathcal{C}_k) - \log p_{\mathbf{x}}(\bm{x}) \\
    &= - \frac{1}{2} [\bm{x} - \bm{\mu}_k]^\top \bm{\Sigma}_k^{-1} [\bm{x} - \bm{\mu}_k] - \frac{1}{2} \log |\bm{\Sigma}_k| + \log P(\textnormal{y} = \mathcal{C}_k) \\
    &\qquad - \frac{p}{2} \log(2\pi) - \log p_{\mathbf{x}}(\bm{x}) \\
    &= - \frac{1}{2} \bm{x}^\top \bm{\Sigma}_k^{-1} \bm{x} + \bm{x}^\top \bm{\Sigma}_k^{-1} \bm{\mu}_k - \frac{1}{2} \bm{\mu}_k^\top \bm{\Sigma}_k^{-1} \bm{\mu}_k - \frac{1}{2} \log |\bm{\Sigma}_k| + \log P(\textnormal{y} = \mathcal{C}_k) \\
    &\qquad - \frac{p}{2} \log(2\pi) - \log p_{\mathbf{x}}(\bm{x})
    \label{eq:log_probability}
\end{split}
\end{align}

\begin{figure}
    \centering
    \includegraphics[width=\textwidth]{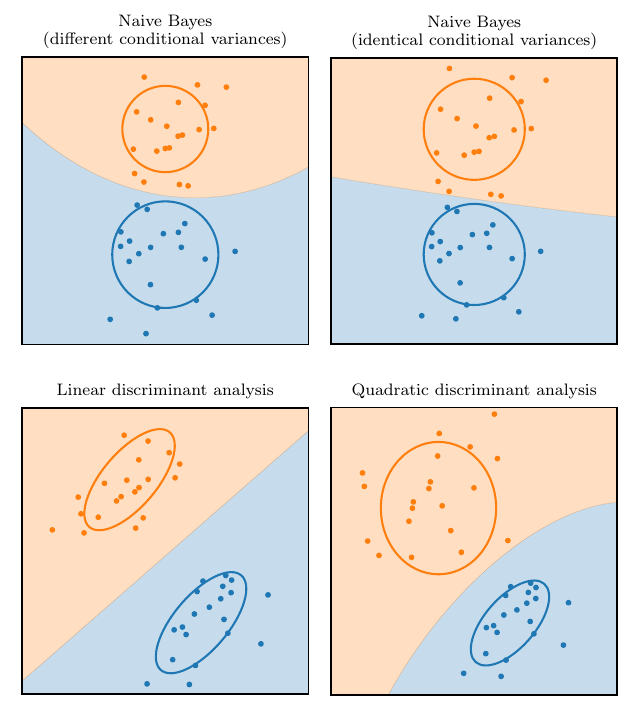}
    \caption{
    Illustration of decision functions with normal distributions.
    A two-dimensional covariance matrix can be represented as an ellipse.
    In the Naive Bayes model, the features are assumed to be independent and to have the same variance conditionally to the class, leading to covariance matrices being represented as circles.
    When the covariance matrices are assumed to be identical, the decision functions are linear instead of quadratic.
    }
    \label{fig:decision_functions_normal_distributions}
\end{figure}

It is also possible to make further assumptions on the covariance matrices that lead to different models.
In this section, we present the most commonly used ones: Naive Bayes, linear discriminant analysis and quadratic discriminant analysis.
\autoref{fig:decision_functions_normal_distributions} illustrates the covariance matrices and the decision functions for these models in the two-dimensional case.

\subsection{Naive Bayes}

The Naive Bayes model assumes that, conditionally to each class $\mathcal{C}_k$, the features are independent and have the same variance $\sigma_k^2$:
$$
\forall k,\, \bm{\Sigma}_k = \sigma_k^2 \bm{I}_p
$$
\autoref{eq:log_probability} can thus be further simplified:
\begin{align*}
    &\log P(\textnormal{y} = \mathcal{C}_k | \mathbf{x} = \bm{x}) \\
    &=  - \frac{1}{2 \sigma_k^2} \bm{x}^\top \bm{x} + \frac{1}{\sigma_k^2} \bm{x}^\top \bm{\mu}_k - \frac{1}{2 \sigma_k^2} \bm{\mu}_k^\top \bm{\mu}_k - \log \sigma_k + \log P(\textnormal{y} = \mathcal{C}_k) \\
    &\qquad - \frac{p}{2} \log(2\pi) - \log p_{\mathbf{x}}(\bm{x}) \\
    &= \bm{x}^\top \bm{W}_k \bm{x} + \bm{x}^\top \bm{w}_k + w_{0k} + s
\end{align*}
where:
\begin{itemize}
    \item $ \bm{W}_k = - \frac{1}{2 \sigma_k^2} \bm{I}_p $ is the matrix of the quadratic term for class $\mathcal{C}_k$,
    \item $\bm{w}_k = \frac{1}{\sigma_k^2} \bm{\mu}_k$ is the vector of the linear term for class $\mathcal{C}_k$,
    \item $w_{0k} = - \frac{1}{2 \sigma_k^2} \bm{\mu}_k^\top \bm{\mu}_k - \log \sigma_k + \log P(\textnormal{y} = \mathcal{C}_k)$ is the intercept for class $\mathcal{C}_k$, and
    \item $s = - \frac{p}{2} \log(2\pi) - \log p_{\mathbf{x}}(\bm{x})$ is a term that does not depend on class $\mathcal{C}_k$.
\end{itemize}
Therefore, Naive Bayes is a quadratic model.
The probabilities for input $\bm{x}$ to belong to each class $\mathcal{C}_k$ can then easily be computed:
$$
P(\textnormal{y} = \mathcal{C}_k | \mathbf{x} = \bm{x}) = \frac{ \exp\left( \bm{x}^\top \bm{W}_k \bm{x} + \bm{x}^\top \bm{w}_k + w_{0k} \right) }{ \sum_{j=1}^k \exp\left( \bm{x}^\top \bm{W}_j \bm{x} + \bm{x}^\top \bm{w}_j + w_{0j} \right) }
$$

With the Naive Bayes model, it is relatively common to have the conditional variances $\sigma_k^2$ to all be equal:
$$
\forall k, \bm{\Sigma}_k = \sigma_k^2 \bm{I}_p = \sigma^2 \bm{I}_p
$$
In this case, \autoref{eq:log_probability} can be even further simplified:
\begin{align*}
    &\log P(\textnormal{y} = \mathcal{C}_k | \mathbf{x} = \bm{x}) \\
    &=  - \frac{1}{2 \sigma^2} \bm{x}^\top \bm{x} + \frac{1}{\sigma^2} \bm{x}^\top \bm{\mu}_k - \frac{1}{2 \sigma^2} \bm{\mu}_k^\top \bm{\mu}_k - \log \sigma_k + \log P(\textnormal{y} = \mathcal{C}_k) \\
    &\qquad - \frac{p}{2} \log(2\pi) - \log p_{\mathbf{x}}(\bm{x}) \\
    &= \bm{x}^\top \bm{w}_k + w_{0k} + s
\end{align*}
where:
\begin{itemize}
    \item $\bm{w}_k = \frac{1}{\sigma^2} \bm{\mu}_k$ is the vector of the linear term for class $\mathcal{C}_k$,
    \item $w_{0k} = - \frac{1}{2 \sigma^2} \bm{\mu}_k^\top \bm{\mu}_k + \log P(\textnormal{y} = \mathcal{C}_k)$ is the intercept for class $\mathcal{C}_k$, and
    \item $s = - \frac{1}{2 \sigma^2} \bm{x}^\top \bm{x} - \log \sigma - \frac{p}{2} \log(2\pi) - \log p_{\mathbf{x}}(\bm{x})$ is a term that does not depend on class $\mathcal{C}_k$.
\end{itemize}
In this case, Naive Bayes becomes a linear model.

\subsection{Linear discriminant analysis}

Linear discriminant analysis (LDA) makes the assumption that all the covariance matrices are identical but otherwise arbitrary:
$$
\forall k,\, \bm{\Sigma}_k = \bm{\Sigma}
$$
Therefore, \autoref{eq:log_probability} can be further simplified:
\begin{align*}
    &\log P(\textnormal{y} = \mathcal{C}_k | \mathbf{x} = \bm{x}) \\
    &= - \frac{1}{2} [\bm{x} - \bm{\mu}_k]^\top \bm{\Sigma}^{-1} [\bm{x} - \bm{\mu}_k] - \frac{1}{2} \log |\bm{\Sigma}| + \log P(\textnormal{y} = \mathcal{C}_k) \\
    &\qquad - \frac{p}{2} \log(2\pi) - \log p_{\mathbf{x}}(\bm{x}) \\
    &= - \frac{1}{2} \left( \bm{x}^\top \bm{\Sigma}^{-1} \bm{x} - \bm{x}^\top \bm{\Sigma}^{-1} \bm{\mu}_k - \bm{\mu}_k^\top \bm{\Sigma}^{-1} \bm{x} + \bm{\mu}_k^\top \bm{\Sigma}^{-1} \bm{\mu}_k \right) \\
    &\qquad - \frac{1}{2} \log |\bm{\Sigma}| + \log P(\textnormal{y} = \mathcal{C}_k) - \frac{p}{2} \log(2\pi) - \log p_{\mathbf{x}}(\bm{x}) \\
    &= - \bm{x}^\top \bm{\Sigma}^{-1} \bm{\mu}_k - \frac{1}{2} \bm{x}^\top \bm{\Sigma}^{-1} \bm{x} - \frac{1}{2} \bm{\mu}_k^\top \bm{\Sigma}^{-1} \bm{\mu}_k  + \log P(\textnormal{y} = \mathcal{C}_k) - \frac{1}{2} \log |\bm{\Sigma}| \\ 
    &\qquad - \frac{p}{2} \log(2\pi) - \log p_{\mathbf{x}}(\bm{x}) \\
    &= \bm{x}^\top \bm{w}_k + w_{0k} + s
\end{align*}
where:
\begin{itemize}
    \item $\bm{w}_k = \bm{\Sigma}^{-1} \bm{\mu}_k$ is the vector of coefficients for class $\mathcal{C}_k$,
    \item $w_{0k} = - \frac{1}{2} \bm{\mu}_k^\top \bm{\Sigma}^{-1} \bm{\mu}_k + \log P(\textnormal{y} = \mathcal{C}_k)$ is the intercept for class $\mathcal{C}_k$, and
    \item $s = - \frac{1}{2} \bm{x}^\top \bm{\Sigma}^{-1} \bm{x} - - \frac{1}{2} \log |\bm{\Sigma}| - \frac{p}{2} \log(2\pi) - \log p_{\mathbf{x}}(\bm{x})$ is a term that does not depend on class $\mathcal{C}_k$.
\end{itemize}
Therefore, linear discriminant analysis is a linear model.
When $\bm{\Sigma}$ is diagonal, linear discriminant analysis is identical to Naive Bayes with identical conditional variances.

The probabilities for input $\bm{x}$ to belong to each class $\mathcal{C}_k$ can then easily be computed:
$$
P(\textnormal{y} = \mathcal{C}_k | \mathbf{x} = \bm{x})
= \frac{ \exp\left( \bm{x}^\top \bm{w}_k + w_{0k} \right) }{ \sum_{j=1}^k \exp\left( \bm{x}^\top \bm{w}_j + w_{0j} \right) }
$$

\subsection{Quadratic discriminant analysis}

Quadratic discriminant analysis makes no assumption on the covariance matrices $\bm{\Sigma}_k$ that can all be arbitrary.
\autoref{eq:log_probability} can be written as:
\begin{align*}
    &\log P(\textnormal{y} = \mathcal{C}_k | \mathbf{x} = \bm{x}) \\
    &= - \frac{1}{2} \bm{x}^\top \bm{\Sigma}_k^{-1} \bm{x} + \bm{x}^\top \bm{\Sigma}_k^{-1} \bm{\mu}_k - \frac{1}{2} \bm{\mu}_k^\top \bm{\Sigma}_k^{-1} \bm{\mu}_k - \frac{1}{2} \log |\bm{\Sigma}_k| + \log P(\textnormal{y} = \mathcal{C}_k) \\
    &\qquad - \frac{p}{2} \log(2\pi) - \log p_{\mathbf{x}}(\bm{x}) \\
    &= \bm{x}^\top \bm{W}_k \bm{x} + \bm{x}^\top \bm{w}_k + w_{0k} + s
\end{align*}
where:
\begin{itemize}
    \item $ \bm{W}_k = - \frac{1}{2} \bm{\Sigma}_k^{-1} $ is the matrix of the quadratic term for class $\mathcal{C}_k$,
    \item $\bm{w}_k = \bm{\Sigma}_k^{-1} \bm{\mu}_k$ is the vector of the linear term for class $\mathcal{C}_k$,
    \item $w_{0k} = - \frac{1}{2} \bm{\mu}_k^\top \bm{\Sigma}_k^{-1} \bm{\mu}_k - \frac{1}{2} \log |\bm{\Sigma}_k| + \log P(\textnormal{y} = \mathcal{C}_k)$ is the intercept for class $\mathcal{C}_k$, and
    \item $s = - \frac{p}{2} \log(2\pi) - \log p_{\mathbf{x}}(\bm{x})$ is a term that does not depend on class $\mathcal{C}_k$.
\end{itemize}
Therefore, quadratic discriminant analysis is a quadratic model.

The probabilities for input $\bm{x}$ to belong to each class $\mathcal{C}_k$ can then easily be computed:
$$
P(\textnormal{y} = \mathcal{C}_k | \mathbf{x} = \bm{x}) = \frac{ \exp\left( \bm{x}^\top \bm{W}_k \bm{x} + \bm{x}^\top \bm{w}_k + w_{0k} \right) }{ \sum_{j=1}^k \exp\left( \bm{x}^\top \bm{W}_j \bm{x} + \bm{x}^\top \bm{w}_j + w_{0j} \right) }
$$

\section{Tree-based methods}
\label{sec:tree}
\subsection{Decision tree}

\begin{figure}
    \centering
    \begin{tikzpicture}[
    sibling distance        = 22em,
    level distance          = 12em,
    every node/.style       = {shape=rectangle},
    edge from parent/.style = {draw, -latex},
    sloped,
    treenode/.style         = {shape=rectangle, draw, align=center, inner sep=4},
    yellownode/.style       = {treenode, fill=yellow!30},
    orangenode/.style       = {treenode, fill=orange!30},
    rednode/.style          = {treenode, fill=red!30},
    scale                   = 0.5
]

\node [treenode] {Severity of symptoms}
    child { node [yellownode] {Consult your\\ general practitioner}
      edge from parent node [above, font=\footnotesize] {Mild} }
    child { node [treenode] {Can you quickly\\ consult a specialist?}
      child { node [orangenode] {Consult a specialist}
        edge from parent node [above, font=\footnotesize] {Yes} }
      child { node [rednode] {Call for\\ emergency services}
        edge from parent node [above, font=\footnotesize] {No} }
    edge from parent node [above, font=\footnotesize] {Severe}};

\end{tikzpicture}
    \caption{
    A general thought process when being ill.
    Depending on conditional statements (severity of symptoms, ability to quickly consult a specialist), the decision (consult your general practitioner or a specialist, or call for emergency services) is different.
    }
\label{fig:decision_tree_everyday_life}
\end{figure}
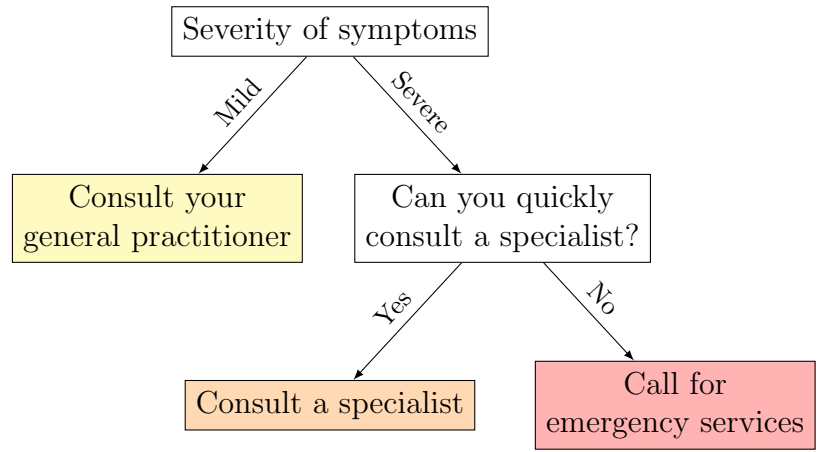

Binary decisions based on conditional statements are frequently used in everyday life because they are intuitive and easy to understand.
\autoref{fig:decision_tree_everyday_life} illustrates a general approach when someone is ill.
Depending on conditional statements (severity of symptoms, ability to quickly consult a specialist), the decision (consult your general practitioner or a specialist, or call for emergency services) is different.
Models with such an architecture are often used in machine learning and are called \emph{decision trees}.

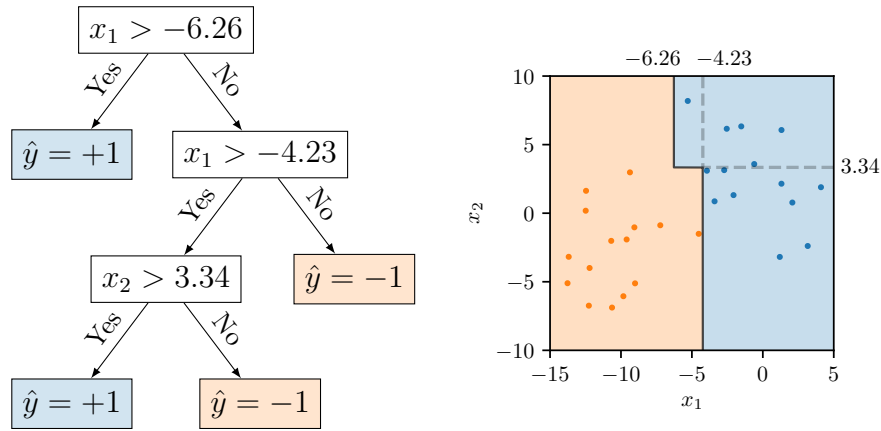
\begin{figure}
\centering
    \begin{subfigure}{0.45\textwidth} \centering
        \begin{tikzpicture}[
    sibling distance        = 12em,
    level distance          = 8em,
    every node/.style       = {shape=rectangle},
    edge from parent/.style = {draw, -latex},
    sloped,
    treenode/.style         = {shape=rectangle, draw, align=center, inner sep=4},
    positivenode/.style     = {treenode, fill=C0!20},
    negativenode/.style     = {treenode, fill=C1!20},
    scale                   = 0.5
]

\node [treenode] {$x_1 > -6.26$}
    child { node [positivenode] {$\hat{y} = +1$}
      edge from parent node [above, font=\footnotesize] {Yes} }
    child { node [treenode] {$x_1 > -4.23$}
      child { node [treenode] {$x_2 > 3.34$}
        child { node [positivenode] {$\hat{y} = +1$}
          edge from parent node [above, font=\footnotesize] {Yes} }
        child { node [negativenode] {$\hat{y} = -1$}
            edge from parent node [above, font=\footnotesize] {No} }
        edge from parent node [above, font=\footnotesize] {Yes} }
      child { node [negativenode] {$\hat{y} = -1$}
        edge from parent node [above, font=\footnotesize] {No}}
    edge from parent node [above, font=\footnotesize] {No}};
\end{tikzpicture}
    \end{subfigure}
    \begin{subfigure}{0.45\textwidth} \centering
        \includegraphics[width=1.\textwidth]{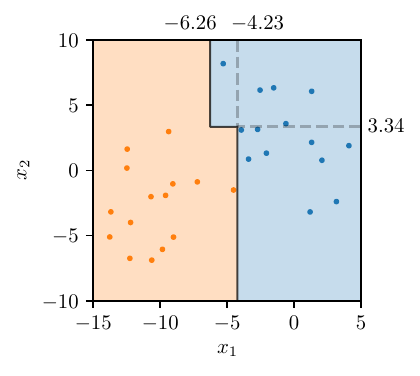}
    \end{subfigure}
    \caption[A decision tree]
    {A decision tree: (left) the rules learned by the decision tree, and (right) the corresponding decision function.}
\label{fig:classification_decision_tree}
\end{figure}

A decision tree is an algorithm containing only conditional statements and can be represented with a tree \citep{breiman_classification_1984}.
This graph consists of:
\begin{itemize}
    \item decision nodes for all the conditional statements,
    \item branches for the potential outcomes of each decision node, and
    \item leaf nodes for the final decision.
\end{itemize}
\autoref{fig:classification_decision_tree} illustrates a decision tree and its corresponding decision function.
For a given sample, the final decision is obtained by following its corresponding path, starting at the root node.

A decision tree recursively partitions the feature space in order to group samples with the same labels or similar target values.
At each node, the objective is to find the best (feature, threshold) pair so that both subsets obtained with this split are the most \emph{pure}, that is homogeneous.
To do so, the best (feature, threshold) pair is defined as the pair that minimizes an \emph{impurity} criterion.

\begin{figure}
    \centering
    \includegraphics[width=1.\textwidth]{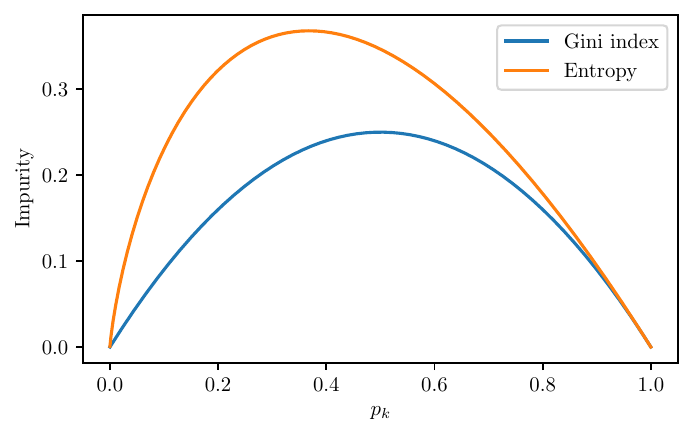}
    \caption{
    Illustration of Gini index and entropy.
    The entropy function takes larger values than the Gini index, especially for $p_k < 0.8$, thus is more discriminative against heterogeneous subsets (when most classes only represent only a small proportion of the samples) than Gini index.
    }
    \label{fig:decision_tree_impurity_criteria}
\end{figure}

Let $\mathcal{S} \subseteq \mathcal{X}$ be a subset of training samples.
For classification tasks, the distribution of the classes, that is the proportion of each class, is used to measure the purity of the subset.
Let $p_{k}$ be the proportion of samples from class $\mathcal{C}_k$ in a given partition:
$$
p_k = \frac{1}{\vert \mathcal{S} \vert} \sum_{y \in \mathcal{S}} \bm{1}_{y = \mathcal{C}_k}
$$
Popular impurity criterion for classification tasks include:
\begin{itemize}
    \item Gini index: $\displaystyle \sum_{k} p_k (1 - p_k)$
    \item Entropy: $\displaystyle - \sum_{k} p_k \log(p_k)$
    \item Misclassification: $\displaystyle 1 - \max_{k} p_k$
\end{itemize}
\autoref{fig:decision_tree_impurity_criteria} illustrates the values of the Gini index and the entropy for a single class $\mathcal{C}_k$ and for different proportions of samples $p_k$.
One can see that the entropy function takes larger values than the Gini index, especially for $p_k < 0.8$.
Since the sum of the proportions is equal to $1$, most classes only represent a small proportion of the samples.
Therefore, a simple interpretation is that entropy is more discriminative against heterogeneous subsets than the Gini index.
Misclassification only takes into account the proportion of the most common class and tends to be even less discriminative against heterogeneous subsets than both entropy and Gini index.

For regression tasks, the mean error from a reference value (such as the mean or the median) is often used as the impurity criterion:
\begin{itemize}
    \item Mean squared error: $\displaystyle \frac{1}{\vert \mathcal{S} \vert} \sum_{y \in \mathcal{S}} (y - \bar{y})^2 \quad \text{with} \quad \bar{y} = \frac{1}{\vert \mathcal{S} \vert}  \sum_{y \in \mathcal{S}} y$
    \item Mean absolute error: $\displaystyle \frac{1}{\vert \mathcal{S} \vert} \sum_{y \in \mathcal{S}} \vert y - \text{median}_{\mathcal{S}}(y) \vert$
\end{itemize}

Theoretically, a tree can grow until every leaf node is perfectly pure.
However, such a tree would have a lot of branches and would be very complex, making it prone to overfitting.
Several strategies are commonly used to limit the size of the tree.
One approach consists in growing the tree with no restriction, then \emph{pruning} the tree, that is replacing subtrees with nodes \citep{breiman_classification_1984}.
Other popular strategies to limit the complexity of the tree are usually applied while the tree is grown, and include setting:
\begin{itemize}
    \item a maximum depth for the tree,
    \item a minimum number of samples required to be at an internal node,
    \item a minimum number of samples required to split a given partition,
    \item a maximum number of leaf nodes,
    \item a maximum number of features considered (instead of all the features) to find the best split,
    \item a minimum impurity decrease to split an internal node.
\end{itemize}

\subsection{Random forest}

One limitation of decision trees is their simplicity.
Decision trees tend to use a small fraction of the features in their decision function.
In order to use more features in the decision tree, growing a larger tree is required, but large trees tend to suffer from overfitting, that is having a low bias but a high variance.
One solution to decrease the variance without much increasing the bias is to build an ensemble of trees with randomness, hence the name \emph{random forest} \citep{breiman_random_2001}.

In a bid to have trees that are not perfectly correlated (thus building actually different trees), each tree is built using only a subset of the training samples obtained with random sampling.
Moreover, for each decision node of each tree, only a subset of the features are considered to find the best split.

The final prediction is obtained by averaging the predictions of each tree.
For classification tasks, the predicted class is either the most commonly predicted class (hard-voting) or the one with the highest mean probability estimate (soft-voting) across the trees.
For regression tasks, the predicted value is usually the mean of the predicted values across the trees.

\begin{floatbox}[t]
    \begin{nicebox}[Random forest]
        \label{box:random_forest}
        \begin{itemize}[leftmargin=2mm]
            \item \textbf{Random forest}: ensemble of decision trees with randomness introduced to build different trees.
            \item \textbf{Decision tree}: algorithm containing only conditional statements and represented with a tree.
            \item \textbf{Regularization}: maximum depth for each tree, minimum number of samples required to split a given partition, etc.
        \end{itemize}
    \end{nicebox}
\end{floatbox}

\subsection{Extremely randomized trees}

Even though random forests involve randomness in sampling both the samples and the features, trees inside a random forest tend to be correlated, thus limiting the variance decrease.
In order to decrease even more the variance of the model (while possibly increasing its bias) by growing less correlated trees, \emph{extremely randomized trees} introduce more randomness \citep{geurts_extremely_2006}.
Instead of looking for the best split among all the candidate (feature, threshold) pairs, one threshold is drawn at random for each candidate feature and the best of these randomly generated thresholds is chosen as the splitting rule.

\section{Clustering}
\label{sec:clustering}
So far, we have presented classic machine learning methods for classification and regression, which are the main components of supervised learning.
Each input $\bm{x}^{(i)}$ had an associated output $y^{(i)}$.
In this section we present clustering, which is an unsupervised machine learning task.
In unsupervised learning, only the inputs $\bm{x}^{(i)}$ are available, with no associated outputs.
As the ground truth is not available, the objective is to extract information from the input data without supervising the learning process with the output data.

Clustering consists in finding groups of samples such that:
\begin{itemize}
    \item samples from the same group are similar, and
    \item samples from different groups are different.
\end{itemize}
For instance, clustering can be used to identify disease subtypes for heterogeneous diseases such as Alzheimer's disease and Parkinson's disease.

In this section, we present two of the most common clustering methods: the $k$-means algorithm and the Gaussian mixture model.

\subsection{\texorpdfstring{$\bm{k}$}{k}-means}

The $k$-means algorithm divides a set of $n$ samples, denoted by $\mathcal{X}$, into a set of $k$ disjoint clusters, each denoted by $\mathcal{X}_j$, such that $\mathcal{X} = \{\mathcal{X}_1, \ldots, \mathcal{X}_k\}$.

\begin{figure}
    \centering
    \includegraphics[width=1.\textwidth]{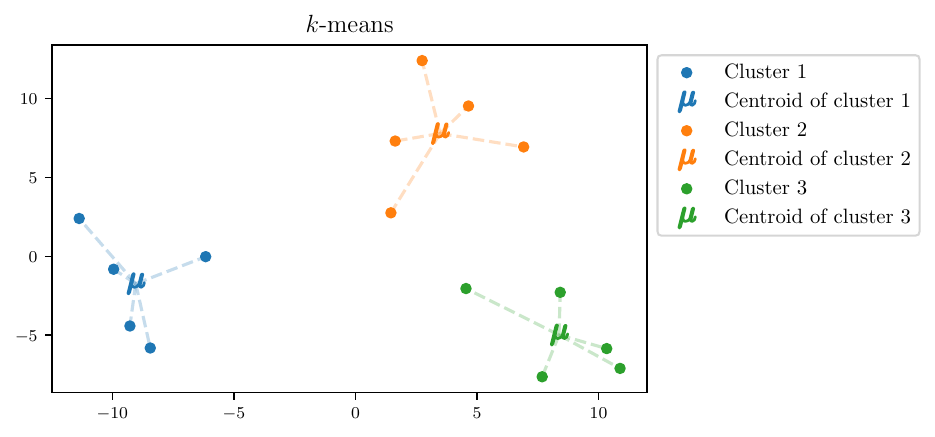}
    \caption{Illustration of the $k$-means algorithm.
    The objective of the algorithm is to find the centroids that minimize the within-cluster sum-of-squares criterion.
    In this example, the inertia is approximately equal to 184.80 and is the lowest possible inertia, meaning that the represented centroids are optimal.}
    \label{fig:kmeans}
\end{figure}

\autoref{fig:kmeans} illustrates the concept of this algorithm.
Each cluster $\mathcal{X}_j$ is characterized by its \emph{centroid}, denoted by $\bm{\mu}_j$, that is the mean of the samples in this cluster:
$$
\bm{\mu}_j = \frac{1}{\vert \mathcal{X}_j \vert} \sum_{\bm{x}^{(i)} \in \mathcal{X}_j} \bm{x}^{(i)}
$$
The centroids fully define the set of clusters because each sample is assigned to the cluster whose centroid is the closest.

The $k$-means algorithm aims at finding centroids that minimize the \emph{inertia}, also known as \emph{within-cluster sum-of-squares criterion}:
$$
\min_{\{\bm{\mu}_1, \ldots, \bm{\mu}_k\}} \sum_{j=1}^k \sum_{\bm{x}^{(i)} \in \mathcal{X}_j} \Vert \bm{x}^{(i)} - \bm{\mu}_j \Vert_2^2
$$
The original algorithm used to find the centroids is often referred to as the \emph{Lloyd's algorithm} \citep{lloyd_least_1982} and is presented in \autoref{alg:kmeans}.
After initializing the centroids, a two-step loop is repeated until convergence (when the centroids are identical for two consecutive iterations) consisting of:
\begin{enumerate}
    \item the \emph{assignment step}, where the clusters are updated based on the current centroids, and
    \item the \emph{update step}, where the centroids are updated based on the current clusters.
\end{enumerate}
When clusters are well-defined, a point from a given cluster is likely to stay in this cluster.
Therefore, the assignment step can be sped up thanks to the triangle inequality by keeping track of lower and upper bounds for distances between points and centers, at the cost of higher memory usage \citep{elkan_using_2003}. 

\begin{algorithm}[t]

\BlankLine

\KwResult{Centroids $\{\bm{\mu}_1, \ldots, \bm{\mu}_k \}$}

\BlankLine

Initialize the centroids $\{\bm{\mu}_1, \ldots, \bm{\mu}_k \}$ \;
    
\BlankLine

 \While{not converged}{

  \BlankLine
 
  \textbf{Assignment step}: Compute the clusters (i.e., assign each sample to its nearest centroid): 
  $$
  \forall j \in \{1, \ldots, k\}, \; \mathcal{X}_j = \{ \bm{x}^{(i)} \in \mathcal{X} \, \vert \, \Vert \bm{x}^{(i)} - \bm{\mu}_j \Vert_2^2 = \min_l \Vert \bm{x}^{(i)} - \bm{\mu}_l \Vert_2^2 \}
  $$
  \\
  \textbf{Update step}: Compute the centroids of the updated clusters:
  $$
  \forall j \in \{1, \ldots, k\}, \; \bm{\mu}_j = \frac{1}{\vert \mathcal{X}_j \vert} \sum_{\bm{x}^{(i)} \in \mathcal{X}_j} \bm{x}^{(i)}
  $$
 }
 \caption{Lloyd's algorithm (a.k.a. naive $k$-means algorithm).}
 \label{alg:kmeans}
\end{algorithm}

Even though the $k$-means algorithm is one of the simplest and most used clustering methods, it has several downsides that should be kept in mind.

First, the number of clusters $k$ is a hyperparameter.
Setting a value much different from the actual number of clusters may yield poor clusters.

\begin{figure}
    \centering
    \includegraphics[width=1.\textwidth]{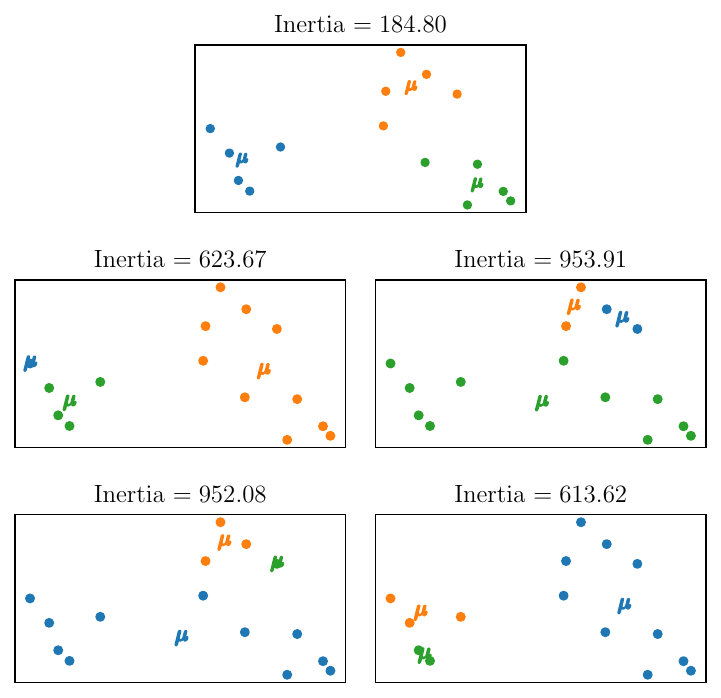}
    \caption{Illustration of the convergence of the $k$-means algorithm to bad local minima.
    In the upper figure, the algorithm converged to a global minimum because the inertia is equal to the minimum possible value (184.80), thus the obtained clusters are optimal. 
    In the four other figures, the algorithm converged to a local minima that are not global minima because the inertias are higher than the minimum possible value, thus the obtained clusters are suboptimal.}
    \label{fig:kmeans_global_local_minima}
\end{figure}

Second, the inertia is not a convex function.
Although Lloyd's algorithm is guaranteed to converge, it may converge to a local minimum that is not a global minimum.
\autoref{fig:kmeans_global_local_minima} illustrates the convergence to such centroids.
Several strategies are often applied to address this issue, including sophisticated centroid initialization \citep{arthur_k-means_2007} and running the algorithm numerous times and keeping the best run (i.e., the one yielding the lowest inertia).

Third, inertia makes the assumption that the clusters are convex and isotropic.
The $k$-means algorithm may yield poor results when this assumption does not hold, such as with elongated clusters or manifolds with irregular shapes.

Fourth, the Euclidean distance tends to be inflated (i.e., the ratio of the distances of the nearest and farthest neighbors to a given target is close to 1) in high-dimensional spaces, making inertia a poor criterion in such spaces \citep{aggarwal_surprising_2001}.
One can alleviate this issue by running a dimensionality reduction method such as principal component analysis prior to the $k$-means algorithm.

\subsection{Gaussian mixture model}

A mixture model makes the assumption that each sample is generated from a mixture of several independent distributions.

Let $k$ be the number of distributions.
Each distribution $F_j$ is characterized by its probability of being picked, denoted by $\pi_j$, and its density $p_j$ with parameters $\bm{\theta}_j$, denoted by $p_j(\cdot, \bm{\theta}_j)$.
Let $\bm{\Delta} = (\Delta_1, \ldots, \Delta_k)$ be a vector-valued random variable such that:
$$
\sum_{j=1}^k \Delta_j = 1 \quad \text{and} \quad \forall j \in \{1, \ldots, k\}, \, P (\Delta_j = 1) = 1 - P (\Delta_j = 0) = \pi_j
$$
and $(\textnormal{x}_1, \ldots, \textnormal{x}_k)$ be independent random variables such that $\textnormal{x}_j \sim F_j$.
The samples are assumed to be generated from a random variable $\textnormal{x}$ with density $p_{\textnormal{x}}$ such that:
\begin{gather*}
    \textnormal{x} = \sum_{j=1}^k \Delta_j \textnormal{x}_j \\
    \forall \bm{x} \in \mathcal{X}, \, p_{\textnormal{x}}(\bm{x}, \bm{\theta}) = \sum_{j=1}^k \pi_j p_j(\bm{x}; \bm{\theta}_j)
\end{gather*}

\begin{figure}
    \centering
    \includegraphics[width=1.\textwidth]{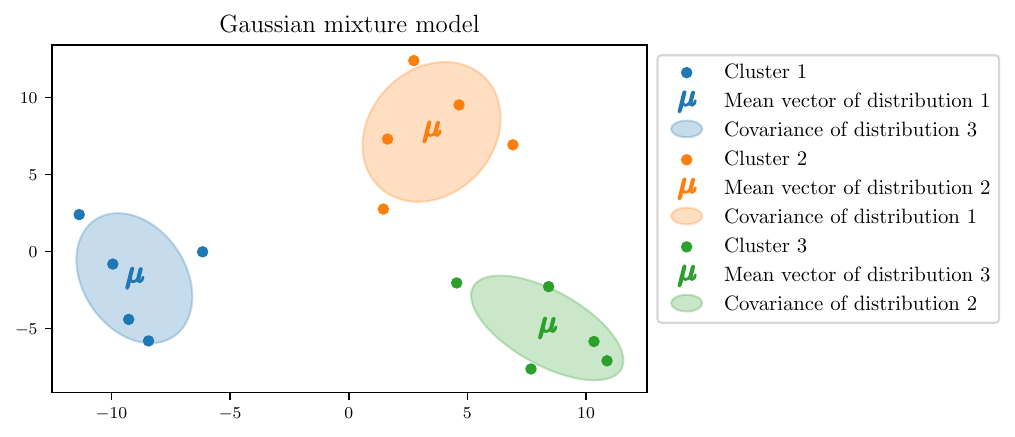}
    \caption{Gaussian mixture model.
    For each estimated distribution, the mean vector and the ellipsis consisting of all the points within one standard deviation of the mean are plotted.}
    \label{fig:gaussian_mixture_model}
\end{figure}

A Gaussian mixture model is a particular case of a mixture model in which each distribution $F_j$ is a Gaussian distribution with mean vector $\bm{\mu}_j$ and covariance matrix $\bm{\Sigma}_j$:
$$
\forall j \in \{1, \ldots, k \}, \, F_j = \mathcal{N}(\bm{\mu}_j, \bm{\Sigma}_j) \\
$$
\autoref{fig:gaussian_mixture_model} illustrates the learned distributions from a Gaussian mixture model.

The objective is to find the parameters $\bm{\theta}$ that maximize the likelihood, with $\bm{\theta} = \left( \{ \bm{\mu}_j \}_{j=1}^k, \{ \bm{\Sigma}_j \}_{j=1}^k, \{ \pi_j \}_{j=1}^k \right)$:
$$
L(\bm{\theta}) = \prod_{i=1}^n p_{X}(\bm{x}^{(i)}; \bm{\theta})
$$
For computational reasons, it is easier to maximize the log-likelihood:
$$
\log(L(\bm{\theta}))
= \sum_{i=1}^n \log(p_{X}(\bm{x}^{(i)}; \bm{\theta}))
= \sum_{i=1}^n \log \left( \sum_{j=1}^k \pi_j p_j(\bm{x}; \bm{\theta}_j) \right)
$$
Because the density $p_{X}(\cdot, \bm{\theta})$ is a weighted sum of Gaussian densities, the expression cannot be further simplified.

In order to solve this maximization problem, an algorithm called \emph{Expectation--Maximization} (EM) is often applied \citep{dempster_maximum_1977}.
\autoref{alg:gaussian_mixture_model} describes the main concepts of this algorithm.
After initializing the parameters of each distribution, a two-step loop is repeated until convergence (when the parameters are stable over consecutive loops):
\begin{itemize}
    \item the \emph{expectation step}, in which the probability for each sample $\bm{x}^{(i)}$ to have been generated from distribution $F_j$ is computed, and
    \item the \emph{maximization step}, in which the probability and the parameters of each distribution are updated.
\end{itemize}
Because it is impossible to know which samples have been generated by each distribution, it is also impossible to directly maximize the log-likelihood, which is why we compute its \emph{expected value} using the posterior probabilities, hence the name \emph{expectation step}.
The second step simply consists in maximizing the expected log-likelihood, hence the name \emph{maximization step}.

\begin{algorithm}[t]

\BlankLine

\KwResult{Mean vectors $\{\bm{\mu}_j\}_{j=1}^k$, covariance matrices $\{\bm{\Sigma}_j\}_{j=1}^k$ and probabilities $\{\pi_j\}_{j=1}^k$}

\BlankLine

Initialize the mean vectors $\{\bm{\mu}_j\}_{j=1}^k$, covariance matrices $\{\bm{\Sigma}_j\}_{j=1}^k$ and probabilities $\{\pi_j\}_{j=1}^k$ \;

\BlankLine

 \While{not converged}{

  \BlankLine

  \textbf{E-step}: Compute the posterior probability $\gamma_i(j)$ for each sample $\bm{x}^{(i)}$ to have been generated from distribution $F_j$: 
  $$
  \forall i \in \{1, \ldots, n \}, \, \forall j \in \{1, \ldots, k\}, \; \gamma_i(j) = \frac{\pi_j p_j(\bm{x}^{(i)}; \bm{\theta}_j, \bm{\Sigma}_j)}{\sum_{l=1}^k \pi_l p_j(\bm{x}^{(i)}; \bm{\theta}_l, \bm{\Sigma}_l)}
  $$
  \\
  \textbf{M-step}: Update the parameters of each distribution $F_j$:
  \begin{align*}
      \forall j \in \{1, \ldots, k\}, \; \bm{\mu}_j &= \frac{\sum_{i=1}^n \gamma_i(j) \bm{x}^{(i)}}{\sum_{i=1}^n \gamma_i(j)} \\
      \forall j \in \{1, \ldots, k\}, \; \bm{\Sigma}_j &= \frac{\sum_{i=1}^n \gamma_i(j) [\bm{x}^{(i)} -\bm{\mu}_j][\bm{x}^{(i)} -\bm{\mu}_j]^\top}{\sum_{i=1}^n \gamma_i(j)} \\
      \forall j \in \{1, \ldots, k\}, \; \pi_j &= \frac{1}{n} \sum_{i=1}^n \gamma_i(j)
  \end{align*}
 }
 \caption{Expectation--Maximization algorithm for Gaussian mixture models.}
 \label{alg:gaussian_mixture_model}
\end{algorithm}

Lloyd's and EM algorithms have a lot of similarities.
In the first step, the assignment step assigns each sample to its closest cluster, whereas the expectation step computes the probability for each sample to have been generated from each distribution.
In the second step, the update step computes the centroid of each cluster as the mean of the samples in a given cluster, while the maximization step updates the probability and the parameters of each distribution as a weighted average over all the samples.
For these reasons, the $k$-means algorithm is often referred to as a \emph{hard-voting} clustering method, as opposed to the Gaussian mixture model which is referred to as a \emph{soft-voting} clustering method.

The Gaussian mixture model has several advantages over the $k$-means algorithm.

First, the use of normal distribution densities instead of Euclidean distances dwindles the inflation issue in high-dimensional spaces. Second, the Gaussian mixture model includes covariance matrices, allowing for clusters with elliptical shapes, while the $k$-means algorithm only include centroids, forcing clusters to have circular shapes.

Nonetheless, the Gaussian mixture model also has several drawbacks, sharing a few with the $k$-means algorithm.

First, the number of distributions $k$ is a hyperparameter.
Setting a value much different from the actual number of clusters may yield poor clusters.
Second, the log-likelihood is not a concave function.
Like Lloyd's algorithm, the EM algorithm is guaranteed to converge but it may converge to a local maximum that is not a global maximum.
Several strategies are often applied to address this issue, including sophisticated centroid initialization \citep{arthur_k-means_2007} and running the algorithm numerous times and keeping the best run (i.e., the one yielding the highest log-likelihood).
Third, the Gaussian mixture model has more parameters than the $k$-means algorithm.
Therefore, it usually requires more samples to accurately estimate its parameters (in particular the covariance matrices) than the $k$-means algorithm.

\section{Dimensionality reduction}
\label{sec:dimensionality_reduction}

Dimensionality reduction consists in finding a good mapping from the input space into a space of lower dimension.
Dimensionality reduction can either be unsupervised or supervised.

\subsection{Principal component analysis}

For exploratory data analysis, it may be interesting to investigate the variances of the $p$ features and the $\frac{1}{2}p (p - 1)$ covariances or correlations.
However, as the value of $p$ increases, this process becomes growingly tedious.
Moreover, each feature may explain a small proportion of the total variance.
It may be more desirable to have another representation of the data where a small number of features explain most of the total variance, in other words to have a coordinate system adapted to the input data.

Principal component analysis (PCA) consists in finding a representation of the data through \emph{principal components} \citep{jolliffe_principal_2002}.
The principal components are a sequence of unit vectors such that the $i$-th vector is the best approximation of the data (i.e., maximizing the explained variance) while being orthogonal to the first $i-1$ vectors.

\begin{figure}
    \centering
    \includegraphics[width=0.7\textwidth]{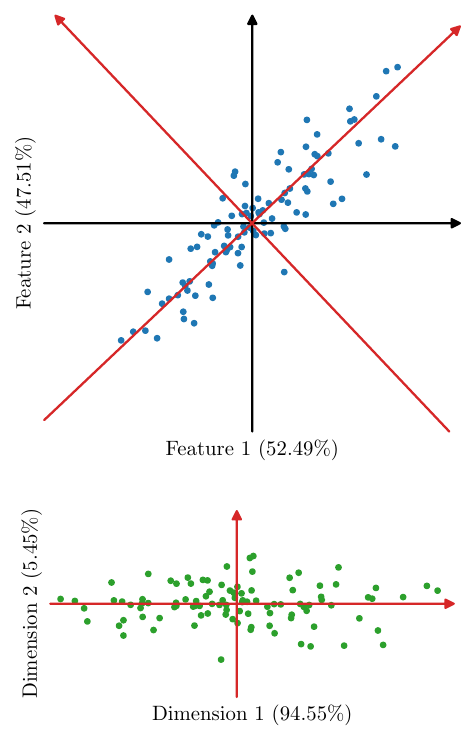}
    \caption{
    Illustration of principal component analysis.
    On the upper figure, the training data in the original space (blue points with black axes) is plotted.
    Both features explain about the same amount of the total variance, although one can clearly see a linear pattern.
    Principal component analysis learns a new Cartesian coordinate system based on the input data (red axes).
    On the lower figure, the training data in the new coordinate system is plotted (green points with red axes).
    The first dimension explains much more variance than the second dimension.
    }
    \label{fig:principal_component_analysis}
\end{figure}

\autoref{fig:principal_component_analysis} illustrates principal component analysis when the input space is two-dimensional.
On the upper figure, the training data in the original space is plotted.
Both features explain about the same amount of the total variance, although one can clearly see that both features are strongly correlated.
Principal component analysis identifies a new Cartesian coordinate system based on the input data.
On the lower figure, the training data in the new coordinate system is plotted.
The first dimension explains much more variance than the second dimension.

\subsubsection{Full decomposition}

Mathematically, given an input matrix $\bm{X} \in \mathbb{R}^{n \times p}$ that is centered (i.e., the mean value of each column $\bm{X}_{:,j}$ is equal to zero), the objective is to find a matrix $\bm{W} \in \mathbb{R}^{p \times p}$ such that:
\begin{itemize}
    \item $\bm{W}$ is an orthogonal matrix, i.e. its columns are unit vectors and orthogonal to each other,
    \item the new representation of the input data, denoted by $\bm{T}$, consists of the coordinates in the Cartesian coordinate system induced by $\bm{W}$ (whose columns form an orthogonal basis of $\mathbb{R}^p$ with the Euclidean dot product):
    $$
    \bm{T} = \bm{X} \bm{W}
    $$
    \item each column of $\bm{W}$ maximizes the explained variance.
\end{itemize}
Each column $\bm{w}_i = \bm{W}_{:,i}$ is a principal component.
Each input vector $\bm{x}$ is transformed into another vector $\bm{t}$ using a linear combination of each feature with the weights from the $\bm{W}$ matrix:
$$
\bm{t} = \bm{x}^\top \bm{W}
$$

The first principal component $\bm{\bm{w}}^{(1)}$ is the unit vector that maximizes the explained variance:
\begin{align*}
    \bm{w}_1 &= \argmax_{\Vert \bm{w} \Vert = 1} \left\{ \sum_{i=1}^n \bm{x}^{(i)\top} \bm{w} \Vert \right\} \\
    &= \argmax_{\Vert \bm{w} \Vert = 1} \left\{ \Vert \bm{X} \bm{w} \Vert \right\} \\
    &= \argmax_{\Vert \bm{w} \Vert = 1} \left\{ \bm{w}^\top \bm{X}^\top \bm{X} \bm{w} \Vert \right\} \\
    \bm{w}_1 &= \argmax_{\bm{w} \in \mathbb{R}^p} \left\{ \frac{\bm{w}^\top \bm{X}^\top \bm{X} \bm{w}}{\bm{w}^\top \bm{w}} \right\} 
\end{align*}
As $\bm{X}^\top \bm{X}$ is a positive semi-definite matrix, a well known result from linear algebra is that $\bm{w}^{(1)}$ is the eigenvector associated with the largest eigenvalue of $\bm{X}^\top \bm{X}$.

The $k$-th component is found by subtracting the first $k - 1$ principal components from $\bm{X}$:
$$
\bm{\hat{X}}_k = \bm{X} - \sum_{s = 1}^{k - 1} \bm{X} \bm{w}^{(s)} \bm{w}^{(s)\top}
$$
and then finding the unit vector that explains the maximum variance from this new data matrix:
$$
\bm{w}_k
= \argmax_{\Vert \bm{w} \Vert = 1} \left\{ \Vert \bm{\hat{X}}_{k} \bm{w} \Vert \right\}
= \argmax_{\bm{w} \in \mathbb{R}^p} \left\{ \frac{\bm{w}^\top \bm{\hat{X}}_{k}^\top \bm{\hat{X}}_{k} \bm{w}}{\bm{w}^\top \bm{w}} \right\}
$$
One can show that the eigenvector associated with the $k$-th largest eigenvalue of the $\bm{X}^\top \bm{X}$ matrix maximizes the quantity to be maximized.

Therefore, the matrix $\bm{W}$ is the matrix whose columns are the eigenvectors of the $\bm{X}^\top \bm{X}$ matrix, sorted by descending order of their associated eigenvalues.

\subsubsection{Truncated decomposition}

Since each principal component iteratively maximizes the remaining variance, the first principal components explain most of the total variance, while the last ones explain a tiny proportion of the total variance.
Therefore, keeping only a subset of the ordered principal components usually gives a good representation of the input data.

\begin{figure}
    \centering
    \includegraphics[width=\textwidth]{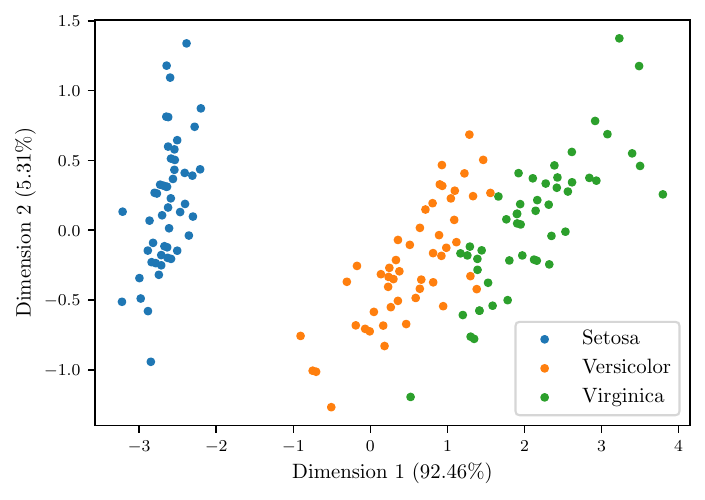}
    \caption{
    Illustration of principal component analysis as a dimensionality reduction technique.
    The Iris flower dataset consists of 50 samples for each of three iris species (setosa, versicolor and virginica) for which four features were measured: the length and the width of the sepals and petals, in centimeters.
    The projection of each sample on the first two principal components is shown in this figure.
    The first dimension explains most of the variance $\mathit{(92.46\%)}$.
    }
    \label{fig:dimensionalty_reduction_pca}
\end{figure}

Mathematically, given a number of dimensions $l$, the new representation is obtained by truncating the matrix of principal components $\bm{W}$ to only keep the first $l$ columns, resulting in the submatrix $\bm{W}_{:,:l}$:
$$
\bm{\tilde{T}} = \bm{X} \bm{W}_{:,:l}
$$
\autoref{fig:dimensionalty_reduction_pca} illustrates the use of principal component analysis as dimensionality reduction.
The Iris flower dataset consists of 50 samples for each of three iris species (setosa, versicolor and virginica) for which four features were measured: the length and the width of the sepals and petals, in centimeters.
The projection of each sample on the first two principal components is shown in this figure.

\subsection{Linear discriminant analysis}

In \autoref{sec:decision_functions_normal_distributions}, we introduced linear discriminant analysis (LDA) as a classification method.
However, it can also be used as a supervised dimensionality reduction method.
LDA fits a multivariate normal distribution for each class $\mathcal{C}_k$, so that each class is characterized by its mean vector $\bm{\mu}_k \in \mathbb{R}^p$ and has the same covariance matrix $\bm{\Sigma} \in \mathbb{R}^{p \times p}$.
However, a set of $k$ points lies in a space of dimension at most $k - 1$.
For instance, a set of 2 points lies on a line, while a set of 3 points lies on a plane.
Therefore, the subspace induced by the $k$ mean vectors $\bm{\mu}_k$ can be used as dimensionality reduction.

There exists another formulation of linear discriminant analysis which is equivalent and more intuitive for dimensionality reduction.
Linear discriminant analysis aims to find a linear projection so that the classes are separated as much as possible (i.e., projections of samples from a same class are close to each other, while projections of samples from different classes are far from each other).

Mathematically, the objective is to find the matrix $\bm{W} \in \mathbb{R}^{p \times l}$ (with $l \leq k - 1$) that maximizes the between-class scatter while also minimizing the within-class scatter:
$$
\max_{\bm{W}} \textnormal{tr}\left( \left( \bm{W}^\top \bm{S}_{w} \bm{W} \right)^{-1} \left( \bm{W}^\top \bm{S}_{b} \bm{W} \right)  \right)
$$
The within-class scatter matrix $\bm{S}_w$ summarizes the diffusion between the mean vector $\bm{\mu}_k$ of class $\mathcal{C}_k$ and all the inputs $\bm{x}^{(i)}$ belonging to class $\mathcal{C}_k$, over all the classes:
$$
\bm{S}_w = \sum_{k=1}^q \sum_{y^{(i)} = \mathcal{C}_k} [\bm{x}^{(i)} - \bm{\mu}_k] [\bm{x}^{(i)} - \bm{\mu}_k]^\top
$$
The between-class scatter matrix $\bm{S}_b$ summarizes the diffusion between all the mean vectors:
$$
\bm{S}_{b} = \sum_{k=1}^q n_k [\bm{\mu}_k - \bm{\mu}] [\bm{\mu}_k - \bm{\mu}]^\top
$$
where $n_k$ is the proportion of samples belonging to class $\mathcal{C}_k$ and $\bm{\mu} = \sum_{k=1}^q n_k \bm{\mu}_k = \frac{1}{n} \sum_{i=1}^n \bm{x}^{(i)} $ is the mean vector over all the input vectors.

One can show that the $\bm{W}$ matrix consists of the first $l$ eigenvectors of the matrix $\bm{S}_w^{-1} \bm{S}_{b}$ with the corresponding eigenvalues being sorted in descending order.
Just as in principal component analysis, the corresponding eigenvalues can be used to determine the contribution of each dimension.
However, the criterion for linear discriminant analysis is different from the one from principal component analysis: it is to maximizing the separability of the classes instead of maximizing the explained variance.

\begin{figure}
    \centering
    \includegraphics[width=\textwidth]{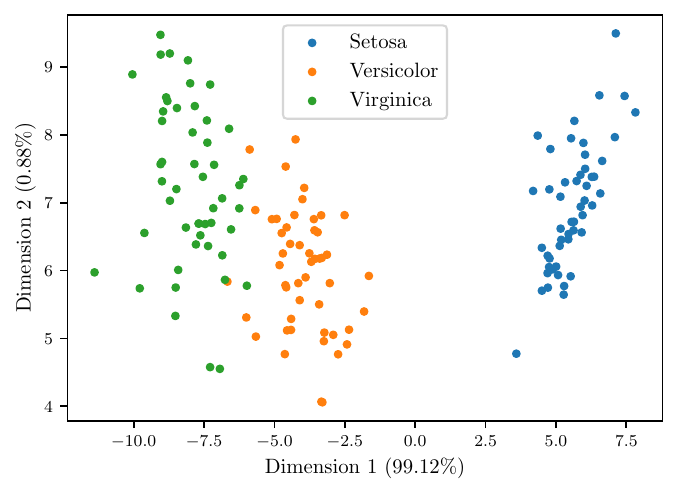}
    \caption{
    Illustration of linear discriminant analysis as a dimensionality reduction technique.
    The Iris flower dataset consists of 50 samples for each of three iris species (setosa, versicolor and virginica) for which four features were measured: the length and the width of the sepals and petals, in centimeters.
    The projection of each sample on the learned two-dimensional space is shown in this figure.
    }
    \label{fig:dimensionalty_reduction_lda}
\end{figure}

\autoref{fig:dimensionalty_reduction_lda} illustrates the use of linear discriminant analysis as a dimensionality reduction technique.
We use the same Iris flower dataset as in \autoref{fig:dimensionalty_reduction_pca} illustrating principal component analysis.
The projection of each sample on the learned two-dimensional space is shown and one can see that the first (horizontal) axis is more discriminative of the three classes with linear discriminant analysis than with principal component analysis.

\section{Kernel methods}
\label{sec:kernel_methods}

Kernel methods allow for generalizing linear models to non-linear models with the use of kernel functions.

As mentioned in \autoref{sec:support_vector_machine}, the main idea of kernel methods is to first map the input data from the original input space to a feature space, and then perform dot products in this feature space.
Under certain assumptions, an optimal solution of the minimization problem of the cost function admits the following form:
$$
f = \sum_{i=1}^n \alpha_i K(\cdot, \bm{x}^{(i)})
$$
where $K$ is the kernel function which is equal to the dot product in the feature space:
$$
\forall \bm{x}, \bm{x}' \in \mathcal{I},\, K(\bm{x}, \bm{x}') = \phi(\bm{x})^\top \phi(\bm{x}')
$$
As this term frequently appears, we denote by $\bm{K}$ the $n \times n$ symmetric matrix consisting of the evaluations of the kernel on all the pairs of training samples:
$$
\forall i, j \in \{1, \ldots, n \},\, K_{ij} = K(\bm{x}^{(i)}, \bm{x}^{(j)})
$$

In this section we present the extension of two models previously introduced in this chapter, ridge regression and principal component analysis, with kernel functions.

\subsection{Kernel ridge regression}

Kernel ridge regression combines ridge regression with the kernel trick, and thus learns a linear function in the space induced by the respective kernel and the training data \cite{murphy_machine_2012}.
For non-linear kernels, this corresponds to a non-linear function in the original input space.

Mathematically, the objective is to find the function $f$ with the following form:
$$
f = \sum_{i=1}^n \alpha_i K(\cdot, \bm{x}^{(i)})
$$
that minimizes the sum of squared errors with a $\ell_2$ penalization term:
$$
\min_{f} \sum_{i=1}^n \left( y^{(i)} - f(\bm{x}^{(i)} \right)^2 + \lambda \Vert f \Vert^2
$$
The cost function can be simplified using the specific form of the possible functions:
\begin{align*}
    &\sum_{i=1}^n (y^{(i)} - f(\bm{x}^{(i)})^2 + \lambda \Vert f \Vert^2 \\
    &= \sum_{i=1}^n \left( y^{(i)} - \sum_{j=1}^n \alpha_j k(\bm{x}^{(j)}, \bm{x}^{(i)}) \right)^2 + \lambda \left\Vert \sum_{i=1}^n \alpha_i K(\cdot, \bm{x}^{(i)}) \right\Vert^2 \\
    &= \sum_{i=1}^n \left( y^{(i)} - \bm{\alpha}^\top \bm{K}_{:, i} \right)^2 + \lambda \bm{\alpha}^\top \bm{K} \bm{\alpha} \\
    &= \Vert \bm{y} - \bm{K} \bm{\alpha} \Vert_2^2 + \lambda \bm{\alpha}^\top \bm{K} \bm{\alpha} \\
\end{align*}
Therefore, the minimization problem is:
$$
\min_{\bm{\alpha}} \Vert \bm{y} - \bm{K} \bm{\alpha} \Vert_2^2 + \lambda \bm{\alpha}^\top \bm{K} \bm{\alpha}
$$
for which a solution is given by:
$$
\bm{\alpha}^\star = \left( \bm{K} + \lambda \bm{I} \right)^{-1} \bm{y}
$$

\autoref{fig:regularization_kernel_ridge_regression} illustrates the prediction function of a kernel ridge regression method with a radial basis function kernel.
The prediction function is non-linear as the kernel is non-linear.

\subsection{Kernel principal component analysis}

As mentioned in \autoref{sec:dimensionality_reduction}, principal component analysis consists in finding the linear orthogonal subspace in the original input space such that each principal component explains the most variance.
The optimal solution is given by the first eigenvectors of $\bm{X}^\top \bm{X}$ with the corresponding eigenvalues being sorted in descending order.

With kernel principal component analysis, the objective is to find the linear orthogonal subspace in the feature space such that each principal component in the feature space explains the most variance \cite{scholkopf_kernel_1999}.
The solution is given by the first $l$ eigenvectors $(\bm{\alpha}_k)_{1 \leq k \leq l}$ of the $\bm{K}$ matrix with the corresponding eigenvalues being sorted in descending order.
The eigenvectors are normalized in order to be unit vectors in the feature space.

\begin{figure}
    \centering
    \includegraphics[width=\textwidth]{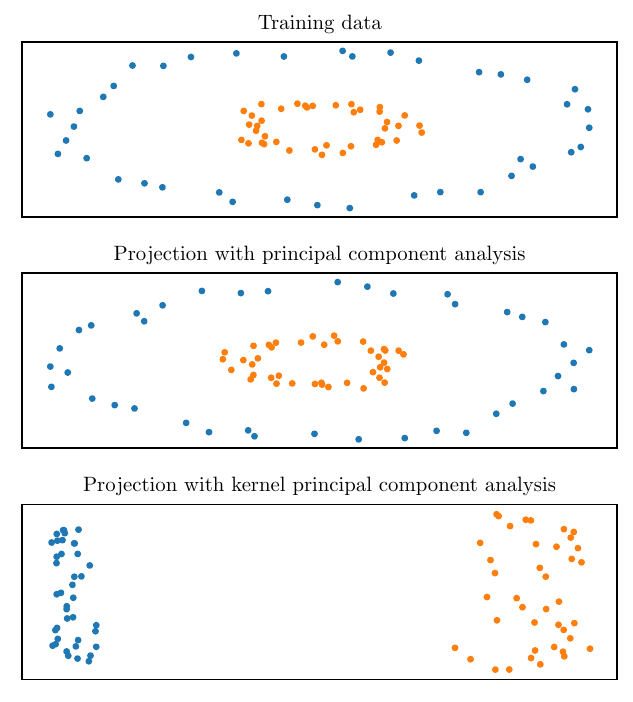}
    \caption{
    Illustration of kernel principal component analysis.
    Some non-linearly separable training data is plotted (top).
    The projected training data using principal component analysis remains non-linearly separable (middle).
    The projected training data using kernel principal component analysis (with a non-linear kernel) becomes linearly separable (bottom).
    }
    \label{fig:kernel_principal_component_analysis}
\end{figure}

Finally, the projection of any input $\bm{x}$ in the original space on the $k$-th component can be computed as:
$$
\phi(\bm{x})^\top \bm{\alpha}_k = \sum_{i=1}^n \alpha_{ki} K(\bm{x}, \bm{x}^{(i)})
$$
\autoref{fig:kernel_principal_component_analysis} illustrates the projection of some non-linearly separable classification data with principal component analysis and with kernel principal component analysis with a non-linear kernel.
The projected input data becomes linearly separable using kernel principal component analysis, whereas the projected input data using (linear) principal component analysis remains non-linearly separable.

\section{Conclusion}
In this chapter, we described the main classic machine learning methods.
Due to space constraints, the description of some of them was brief.
The reader who seeks more details can refer to~\cite{bishop_pattern_2006,hastie_elements_2009}.
All these approaches are implemented in the scikit-learn Python library~\cite{pedregosa2011scikit}.
A common point of the approaches presented in this chapter is that they use as input a set of given or pre-extracted features.
On the contrary, deep learning approaches often provide an end-to-end learning setup within which the features are learned.
These techniques are covered in Chapters 3 to 6.

\section*{Acknowledgments}
The authors would like to thank Hicham Janati for his fruitful remarks.
The authors would like to acknowledge the extensive documentation of the scikit-learn Python package, in particular its user guide, for the relevant information and references provided.
We used the NumPy~\cite{harris2020array}, matplotlib~\cite{hunter2007matplotlib} and scikit-learn~\cite{pedregosa2011scikit} Python packages to generate all the figures.
This work was supported by the French government under management of Agence Nationale de la Recherche as part of the ``Investissements d'avenir'' program, reference ANR-19-P3IA-0001 (PRAIRIE 3IA Institute) and reference ANR-10-IAIHU-06 (Agence Nationale de la Recherche-10-IA Institut Hospitalo-Universitaire-6), and by the European Union H2020 programme (grant number 826421, project TVB-Cloud).

\bibliographystyle{spbasicsort}
\bibliography{references}


\end{document}